# Adaptive Gait Biofeedback With Participant-Held-Out Modeling and Participant-Specific Updating in Chronic Ankle Instability

Jaeyoon (Jason) Kim, Veronika Lebisova, Jeniya Sultana, Jaeyoung Cho, and Jaeho Jang*

***Abstract*—Adaptive gait biofeedback may support repeated practice in chronic ankle instability, but its evaluation must address model performance and human response. We evaluated a temporal convolutional classifier on protocol-defined, angle-derived GOOD/BAD gait-cycle labels using participant-held-out leave-one-subject-out (LOSO) cross-validation in 20 participants. Seven participants in the adaptive-intervention group completed nine sessions over three weeks, with one motion-capture recording analyzed per session. Models updated after failed sessions were compared offline with their parent models on the same-session validation subset used for candidate selection and the first subsequent adaptive-session recording. Frontal-plane ankle angle was compared between the adaptive group and 10 sequentially enrolled controls at Baseline, Post, and 7-day Retention. Across 20 held-out folds, mean fold-level area under the receiver operating characteristic curve (AUROC) was 0.948, sensitivity for angle-threshold-exceeding BAD cycles was 0.941, and specificity for angle-threshold-meeting GOOD cycles was 0.366. Mean BAD-class F1 was higher in candidate models by 0.187 on the same-session subset and 0.118 on the first subsequent recording. At Post, the adaptive group had a baseline-adjusted frontal-plane ankle angle 5.168° lower than controls (95% confidence interval, 1.766°–8.569° lower); the Retention contrast was uncertain. These findings characterize population-model discrimination and offline participant-specific updating during repeated biofeedback use, alongside a nonrandomized Post frontal-plane ankle angle association. They do not establish independent clinical gait classification or a causal benefit of updating.**

## I. Introduction

Chronic ankle instability (CAI) can develop after a lateral ankle sprain and is characterized by recurrent sprains, episodes of giving way, and persistent sensory-perceptual and motor-behavioral impairments [1]. During walking, those with CAI often exhibit greater ankle inversion, including around initial contact and early stance [2],[3],[4]. This altered foot positioning may increase susceptibility to recurrent lateral ankle sprains [1], [3]. Recurrent injury and persistent instability are associated with long-term ankle joint degeneration, including posttraumatic osteoarthritis [5].

Visual gait biofeedback has been used to provide real-time information about ankle positioning during walking in individuals with CAI. In a randomized trial, visual biofeedback targeting ankle inversion at initial contact, combined with impairment-based rehabilitation, reduced inversion at initial contact and throughout the stride cycle [6]. Systematic reviews also report reductions in ankle inversion among selected gait-training and biofeedback interventions [7], [8].

The magnitude of ankle inversion may vary across individuals with CAI and change during gait retraining [4], [6]. A population-trained feedback model may perform differently across these individual gait patterns and as they change over repeated sessions. Participant-specific updating offers a potential way to accommodate these differences, but whether it improves prediction in subsequent sessions requires evaluation. Evidence on biomechanical responses across repeated CAI gait-training sessions and retention after training also remains limited [7], [9], [10]. Evaluating these questions requires a feedback model that can account for sequential gait information across the stride cycle.

Temporal convolutional networks (TCNs) provide one way to map sequential measurements to a feedback state [11]. Related work has examined cross-subject adaptation for wearable-robot intent prediction [12]. Individualized muscle-recruitment prediction has also been used to deliver biofeedback in robotic ankle-exoskeleton training [13]. Because these applications differ in population, sensors, and

Jaeyoon (Jason) Kim and Jeniya Sultana are with the Department of Civil, Environmental, and Construction Engineering, The University of Texas at El Paso, El Paso, TX 79968, USA (e-mail: jkim13@utep.edu; jsultana2@miners.utep.edu).

Jaeyoung Cho is with the Department of Aerospace and Mechanical Engineering, The University of Texas at El Paso, El Paso, TX 79968, USA (e-mail: jcho4@utep.edu).

Veronika Lebisova and Jaeho Jang are with the Department of Kinesiology, The University of Texas at El Paso, El Paso, TX 79968, USA (e-mail: vlebisova@miners.utep.edu; jjang2@utep.edu).

Corresponding author: Jaeho Jang (e-mail: jjang2@utep.edu).

outcomes, their findings do not directly establish model performance in the present CAI workflow.

We developed a TCN-based population classifier for protocol-defined gait-cycle target achievement and participant-specific updating, implemented by fine-tuning after a failed session when another adaptive session remained. GOOD and BAD were operational gait-cycle labels derived from the study's frontal-plane ankle angle rule. The system used model predictions for participant-facing feedback and a separate angle-based rule to determine session success and target progression. The TCN was intended to combine frontal-plane ankle angle with other within-cycle kinematic information when predicting the angle-derived GOOD/BAD label, with the aim of making feedback more responsive to each participant's gait pattern.

Seven participants in the adaptive-intervention (AI) group completed nine adaptive sessions, with one motion-capture recording analyzed from each session. We addressed four questions: (1) How well did the population model distinguish protocol-defined labels in held-out participants? (2) How did participant-specific updating change offline performance on the same-session validation subset used for candidate selection and on the first subsequent adaptive-session recording? (3) How did frontal-plane ankle angle change across repeated adaptive sessions? (4) How did baseline-adjusted frontal-plane ankle angle at Post and Retention differ between the AI and CONTROL cohorts?

The clinical cohorts were enrolled sequentially without random allocation, so group contrasts were interpreted as associations. An accompanying sensitivity analysis assessed dependence on four direct angle-derived input channels under matched training and evaluation conditions. This staged evaluation characterizes how the feedback model performs and what biomechanical responses accompany repeated system use.

## II. Materials and Methods

### A. Study design, participants, and session schedule

The clinical comparison included seven participants in the adaptive-intervention (AI) group and 10 in the no-feedback CONTROL group. CONTROL participants were enrolled first, followed by AI participants, without random allocation. Fig. 1 shows the participant groups and their overlap.

Participants with CAI were recruited using study-specific selection criteria informed by the International Ankle Consortium position statement [14], including (1) a history of at least one significant lateral ankle sprain, (2) recurrent episodes of ankle giving way, (3) an Identification of Functional Ankle Instability (IdFAI) score of ≥11, (4) a Foot and Ankle Ability Measure activities-of-daily-living subscale score of ≤90%, and (5) a Foot and Ankle Ability Measure sport subscale score of ≤80%. Participants were excluded if they reported previous lower-extremity surgery, lower-extremity fracture, or a recent musculoskeletal injury that could affect gait. The involved limb was designated as the tested limb. For bilateral CAI, the limb with the higher IdFAI score was selected for analysis. Institutional review board approval and written informed consent were obtained from all participants.

Between Baseline and Post, AI participants completed nine adaptive biofeedback sessions distributed across three weeks (I01–I09), and CONTROL participants completed nine sessions without feedback. Baseline, Post, and Retention were separate assessment trials from the nine intervention sessions. The Post assessment was conducted on the same day following the final intervention session, with a brief rest break provided when needed. Retention was assessed seven days after the Post assessment.

AI participants received real-time visual biofeedback on a monitor positioned in front of the treadmill. For each completed gait cycle, the TCN-predicted GOOD/BAD class was displayed in green/red, respectively. Participants were instructed to avoid walking on the outside of the foot and to maximize the number of green displays.

The walking-duration and feedback-exposure schedule was adapted from a previously published visual gait-biofeedback protocol [6]. Total walking durations for Sessions I01–I09 were 4, 8, 12, 16, 20, 20, 20, 20, and 20 min, respectively, in both groups. AI participants received visual feedback throughout Sessions I01–I05. During Sessions I06–I09, feedback was provided for 16, 12, 8, and 4 min, respectively, with the remaining walking time completed without feedback. CONTROL participants completed all walking without feedback. This schedule was fixed by session number and was independent of session SUCCESS/FAIL.

Early-stance peak inversion was defined as the maximum inversion angle occurring from initial contact to 100 ms after initial contact. Each participant's baseline early-stance peak inversion value was determined during the 5-min Baseline assessment. The protocol specified nominal target stages at 80%, 60%, 40%, 20%, and 0% of the initial baseline value, beginning at the 80% stage. A session was considered successful when 10 consecutive steps met the recorded angle target, with each step assessed using its measured peak frontal-plane ankle angle within the IC-to-IC + 100-ms event window. The target-meeting step count and session SUCCESS/FAIL status were monitored only by the research team; participants viewed the TCN-predicted green/red feedback. After SUCCESS, the target was reduced for the subsequent session unless it had already reached 0°. If the success criterion was not achieved, the same target was retained for the next session. Once the target reached 0°, no further reductions were applied and the 0° target was maintained for the remaining sessions to avoid excessive eversion.

### B. Gait measurements and model inputs

Three-dimensional marker trajectories were collected using 13 motion-capture cameras (Prime X 13, OptiTrack, NaturalPoint, Corvallis, OR, USA) at 200 Hz while participants walked on an instrumented treadmill (AMTI, Watertown, MA, USA). Ground-reaction-force data were not used in the present analyses. Participants wore standardized running shoes in appropriately fitted sizes throughout testing and intervention sessions. During the Baseline assessment, participants completed treadmill familiarization and a self-selected walking speed was determined. This speed was then used for all subsequent testing and intervention sessions, including the Post and 7-day Retention assessments. Eleven retroreflective markers were placed on the CAI involved limb, including four shank-cluster markers and markers on the first and fifth metatarsals, calcaneus, lateral and medial malleoli, and lateral and medial femoral epicondyles. For Baseline, Post, and Retention testing, participants completed separate 5-min treadmill-walking trials without feedback.

Motion-capture recordings were processed into variable-length gait-cycle sequences. Frontal-plane ankle angle was expressed relative to neutral (0°), with positive values indicating inversion and negative values indicating eversion. Each cycle was represented by a variable-length T × 13 input array, where T was the number of frames in that cycle. We refer to this complete 13-channel representation as Full13. The 0–100% cycle scaling in Fig. 2A was used only for visualization and was not applied to model inputs.

Full13 comprised four channels derived directly from frontal-plane ankle angle: frontal-plane ankle angle (degrees), absolute angle magnitude (degrees), and the first and second time derivatives of frontal-plane ankle angle (degrees/s and degrees/s$^2$). The remaining nine channels were the ankle-to-knee joint-center distance (mm); the first-to-fifth metatarsal marker distance (mm); three ZXY Euler components of foot orientation relative to the shank (degrees); the y-coordinate of the calcaneus marker and the mean y-coordinate of the first and fifth metatarsal markers (mm); and the first time derivatives of those two coordinate signals (mm/s). The y-coordinate channels refer to the recorded coordinate system; alignment with anatomical vertical was not verified. Featurewise z-scores were estimated from event-window frames in each population-model training partition and applied to the corresponding validation and test data. Table S1 provides the input order, detailed calculations, and implementation variable names.

### *C. Cycle labels, feedback, and session progression*

GOOD and BAD were operational gait-cycle labels derived from the study's frontal-plane ankle angle rule; they did not represent independent clinical diagnoses or independently assessed gait quality. For each selected development recording s designated as baseline, the population-label cutoff was cs = 0.8 × the recording-level mean of the per-step peak frontal-plane ankle angles within the initial-contact (IC)-to-IC + 100-ms event window. A retained cycle was labeled BAD (1) if its maximum frontal-plane ankle angle in that event window exceeded cs, and GOOD (0) otherwise. Prior walking studies reported CAI-related ankle differences near initial contact and early stance [3], [6]; the 100-ms boundary was defined by this study's protocol.

For participant-specific fine-tuning, cycle labels were defined using the recorded session target in place of c_s. The cutoff *or* target was not an explicit population-TCN input. For each completed cycle, the TCN estimated the predicted probability of the BAD class, P(BAD), and a probability decision threshold converted this estimate into green/red feedback for the participant.

Session progression was determined separately from the classifier feedback, using the measured angle and recorded target. A target-meeting step had a peak frontal-plane ankle angle at or below the recorded target within the initial-contact (IC)-to-IC + 100-ms event window. Ten consecutive target-meeting steps constituted session SUCCESS. After SUCCESS, the target advanced to a stricter stage, down to a floor of 0°. After FAIL, the recorded target was retained if another adaptive session remained. A failed session with a later adaptive session remaining also triggered participant-specific updating by fine-tuning.

FAIL served as an operational trigger for updating under the hypothesis that a model trained on population data or an earlier session might not match the participant's recorded session target and gait pattern. Relabeling the triggering-session cycles against that target was intended to address this possible mismatch; failure alone does not establish classifier error.

### *D. Population-model training and participant-held-out evaluation*

Population-model development used recordings designated as baseline from 20 participants. These included the 10 CONTROL participants in the clinical comparison and 10 additional participants whose baseline-designated recordings contributed to population-model development. Population-model evaluation assessed classification performance in participants excluded from model fitting. Preprocessing omitted the first 10 s, interpolated short interior marker gaps, and excluded cycles with unresolved event-window gaps, invalid labels, or empty event masks. Each cycle remained a variable-length input sequence, with valid-frame masks used to exclude padding. Detailed segmentation and gap-handling rules are provided in Table S1.

The TCN comprised seven residual blocks with left-padded temporal convolutions, batch normalization, and dropout (Fig. 2). For each cycle, the model averaged features over valid event-window frames and passed them through a classification head to estimate P(BAD). Stored cycles began at or near initial contact, so the theoretical receptive field does not establish use of measured pre-contact history. Detailed architecture, input-derivative calculations, and the scope of the convolutional causality convention are provided in Table S1.

Training used binary cross-entropy with logits and AdamW optimization. Participant-held-out evaluation used leave-one-subject-out (LOSO) cross-validation, with one participant held out for testing in each of 20 folds. In each fold, the remaining 19 participants were divided into 16 for training and three for internal validation. Internal validation determined checkpoint selection, and the test participant contributed neither to normalization nor to training. Training settings are reported in Table S1.

Performance was evaluated with BAD as the positive class and a fixed decision threshold of 0.5. Area under the receiver operating characteristic curve (AUROC) measured ranking performance; BAD-class F1 was the harmonic mean of BAD precision and sensitivity; and balanced accuracy was the mean of BAD sensitivity and GOOD specificity. Each metric was averaged across folds for which it was defined. AUROC, GOOD specificity, and balanced accuracy were undefined for the fold containing only BAD cycles; BAD-class F1 and sensitivity remained defined. Table S2 reports fold-level results and metric availability. A separate model trained on all 20 development participants provided the starting checkpoint for participant-specific fine-tuning. Subsequent updates reused its saved normalization parameters.

### *E. Sensitivity to angle-derived inputs*

To assess how classification performance depended on the four directly angle-derived channels, we defined a reduced nine-channel input representation, termed NoAngle9, by removing frontal-plane ankle angle, absolute angle magnitude, and the first and second time derivatives of frontal-plane ankle angle from Full13. NoAngle9 retained the two distance channels, three foot–shank orientation channels, two marker-coordinate channels, and two coordinate-derivative channels in their original order. These retained channels could still contain information about ankle state. Models trained with Full13 and NoAngle9 were evaluated against the same angle-derived operational labels using the same 20 LOSO folds, training policy, evaluation procedure, and fixed 0.5 decision threshold.

The Full13 model used for this matched training/evaluation comparison was trained separately from the original population model described in Section II-D. Paired fold differences were summarized with 5,000 percentile bootstrap resamples using the same sampled fold indices for both models. Within each resample, the mean was calculated over pairs for which the metric was defined. Table S7 reports the paired comparison, metric availability, and resampling details.

### *F. Participant-specific updating and offline evaluation*

For each failed session with a subsequent adaptive session remaining, recorded cycles were relabeled using the recorded session target and used for participant-specific fine-tuning. Within-session examples were divided chronologically into an initial 80% training partition and a final 20% validation partition. Convolutional weights were held fixed, while the batch-normalization affine parameters and classification head were trainable. Batch normalization was used in training mode, in which its running statistics could update (Table S1).

For each update, the parent checkpoint was the saved pre-update model, and the candidate checkpoint was the saved result of fine-tuning. Parent and candidate checkpoints were compared offline on the same-session validation subset used for candidate selection and on the first subsequent adaptive-session recording, defined as the first analyzed motion-capture recording obtained after the failed session that triggered the update.

The same-session validation subset also informed candidate selection, so its comparison was selection dependent. The first subsequent recording provided temporally separate data, but that comparison was also conducted offline. These analyses therefore evaluated the offline effect of participant-specific updating on saved-model performance rather than live-session performance.

Mean checkpoint differences were computed across update events, so participants with more events contributed proportionally more observations. Uncertainty was summarized with 95% percentile intervals from 3,000 participant-level cluster-bootstrap resamples in which participants were resampled as blocks and all within-participant update events were retained. BAD-class F1 and sensitivity were defined for all 27 events; GOOD specificity and balanced accuracy were defined for 26 because one comparison subset contained no GOOD cycles. Tables S3 and S4 provide event-level comparisons and the sequence of saved checkpoints.

### *G. Repeated adaptive use and subsequent-session response*

The repeated-use analysis included 63 recordings, comprising one analyzed motion-capture recording from each of nine adaptive sessions for each of the seven AI participants. Displays and comparisons used the target recorded for each session. Session SUCCESS/FAIL states and model-update events were summarized across the nine sessions, including I09 failures after which no further update was scheduled. Table S5 reports the analyzed recordings and their recorded targets.

A repeated-measures model with a participant intercept summarized the frontal-plane ankle angle outcome defined in Section II-H as a function of session order. Final-minus-first differences described each participant's change.

To describe human response after sessions that triggered updating, each triggering-session recording was paired with the first subsequent adaptive-session recording. The recorded targets were checked for equality within each pair. An exploratory repeated-event model, fitted using restricted maximum likelihood (REML) with a participant random intercept, estimated the subsequent-minus-triggering frontal-plane ankle angle difference. Given the small repeated-event sample, uncertainty was summarized using both a t-based 95% interval with 6 degrees of freedom and a percentile interval from 3,000 participant-level cluster-bootstrap resamples; further details are provided in Table S5.

### *H. Clinical outcomes and between-group comparison*

For each recording, the clinical frontal-plane ankle angle outcome was the mean of the per-step maximum frontal-plane ankle angles within the initial-contact (IC)-to-IC + 100-ms event window. This recording-level outcome was distinct from the ten-consecutive-step rule used to determine session SUCCESS.

Baseline, Post, and 7-day Retention assessments were obtained from all seven AI and 10 CONTROL participants. For the endpoint analyses, Post was treated as the primary endpoint and Retention as the key secondary endpoint. Separate analysis-of-covariance (ANCOVA) models were fitted for Post and Retention, with group as the between-participant factor and Baseline frontal-plane ankle angle as a covariate. Results are reported as adjusted AI-minus-CONTROL differences with 95% confidence intervals. Robustness was evaluated using the HC3 heteroscedasticity-consistent covariance estimator and participant-level bootstrap confidence intervals; additional sensitivity analyses are reported in Table S6.

## III. Results

### *A. Study sample*

Population-model development included 20 participants; the clinical comparison included seven AI and 10 CONTROL participants, with all CONTROL participants also contributing development recordings (Fig. 1; Table I).

### *B. Population-model performance*

The population model discriminated the operational labels in held-out participants, but recognition of GOOD cycles was limited at the fixed threshold. Across 20 folds and 5,475 cycles, mean fold-level AUROC was 0.948, BAD sensitivity was 0.941, and GOOD specificity was 0.366 (Fig. 3A; Table II).

### C. Dependence on angle-derived inputs

In the matched training/evaluation sensitivity analysis, the model trained with Full13 achieved higher balanced accuracy than the model trained with NoAngle9 (0.690 versus 0.587; paired difference +0.103, 95% percentile interval +0.025 to +0.181). AUROC was also higher with Full13 (0.943 versus 0.866), whereas BAD sensitivity was similar between the two representations (0.921 versus 0.918). Removing the four direct angle-derived channels reduced discrimination, while NoAngle9 retained discrimination against the operational labels. Table S7 reports all paired differences, intervals, and defined-fold counts for this separate analysis.

### D. Offline participant-specific updating

Across 27 updates in seven participants, candidate models had higher mean BAD-class F1 on the same-session validation subset used for candidate selection than their parent models (mean difference +0.187; 95% participant-level cluster-bootstrap interval +0.093 to +0.337). Because this subset also informed candidate selection, the comparison was selection dependent.

On the first subsequent adaptive-session recording, candidates also had higher mean offline BAD-class F1 than their parents (mean difference +0.118; 95% participant-level cluster-bootstrap interval +0.063 to +0.195). These gains in BAD-class F1 and sensitivity were accompanied by lower mean GOOD specificity, while the difference in balanced accuracy on the first subsequent recording remained uncertain. Fig. 3B–C and Table II summarize the offline parent–candidate comparisons.

### E. Repeated use and frontal-plane ankle angle trajectories

The seven AI participants contributed 63 analyzed adaptive-session recordings, nine per participant: 30 SUCCESS and 33 FAIL. Twenty-seven failed sessions with a subsequent adaptive session remaining triggered saved model updates; the six failures in the final adaptive session did not trigger an update.

Frontal-plane ankle angle was lower across successive recordings in the exploratory repeated-measures model (−0.314° per recording; *t*-based 95% interval with 6 degrees of freedom, −0.576° to −0.053°). Five of seven participants had a lower final than first frontal-plane ankle angle (mean final-minus-first difference −2.611°), although individual trajectories were not consistently downward.

The recorded target was unchanged in all 27 pairs of a triggering-session recording and the first subsequent adaptive-session recording. The estimated subsequent-minus-triggering difference was −0.847°, but uncertainty depended on the interval method: the t-based 95% interval with 6 degrees of freedom crossed zero (−2.208° to +0.515°), whereas the participant-level cluster-bootstrap interval did not (−1.459° to −0.134°). Fig. 4 displays the chronological course, equal-target event comparison, and first-to-final summary.

### F. Post and Retention outcomes

At Post, the primary endpoint, AI participants had a lower baseline-adjusted frontal-plane ankle angle than CONTROL participants (AI minus CONTROL −5.168°; 95% CI −8.569° to −1.766°; p = 0.0057). Both robustness intervals also remained below zero (Table S6). At Retention, the key secondary endpoint, the adjusted difference remained uncertain (−2.108°; 95% CI −5.401° to +1.185°; p = 0.1913). Both robustness intervals also included zero (Table S6). Fig. 5 and Table III present individual trajectories, unadjusted group summaries, and adjusted endpoint contrasts.

## IV. Discussion

### A. Principal findings and contribution

This study evaluated a gait-biofeedback workflow combining completed-cycle visual feedback, angle-based target progression, and participant-specific updating. Model predictions provide feedback, while a separate angle-based SUCCESS/FAIL rule governs progression and determines whether a failed session triggers updating when a later adaptive session remains. This distinction links the study’s questions: classification metrics assess agreement with operational labels, session records describe repeated operation, and frontal-plane ankle angle measures the biomechanical response. The findings support repeated workflow operation and show BAD-focused offline model gains alongside a Post frontal-plane ankle angle group difference.

### B. Model performance and operational meaning

Within the development cohort, the population model distinguished operational GOOD/BAD labels in participants excluded from training and fold normalization. High AUROC and BAD sensitivity coexisted with low GOOD specificity at the fixed threshold. For completed-cycle feedback, this means that cycles meeting the operational GOOD definition may still be classified as BAD at the tested threshold. Strong ranking performance therefore does not by itself ensure balanced recognition of both classes.

The meaning of this performance also depends on how the labels were defined. GOOD/BAD labels followed a recording-specific frontal-plane ankle angle cutoff, and the Full13 input representation included the angle trajectory and its derivatives but did not include the cutoff itself. In the separate matched training/evaluation comparison, the reduction in balanced accuracy and AUROC after removal of the four direct angle-derived channels indicates their contribution to discrimination. The discrimination retained with NoAngle9 may reflect ankle-state information in its remaining nine channels. This establishes input dependence within the evaluated system; optimal feature selection, the necessity or superiority of the TCN architecture or temporal context, and independent clinical gait recognition remain unestablished.

### C. Interpretation of participant-specific updating

Updating was motivated by possible differences between prior training data and the participant's current gait pattern and target-defined labels. Saved participant-specific candidates had higher mean offline BAD F1 than their parent checkpoints in both evaluation settings, but mean GOOD specificity was lower and the later balanced-accuracy difference remained uncertain. The pattern therefore supports BAD-focused adjustment to the operational labels. It does not establish that inadequate model feedback caused session failure.

The two evaluation settings provide different levels of separation from model selection. The same-session advantage is conditional on candidate selection because the evaluated subset contributed to selecting the saved candidate. Evaluation on the first subsequent adaptive-session recording extends the comparison beyond the triggering failed session, although it remains an offline comparison. These analyses therefore characterize saved-model performance and do not establish that deployed participant-specific updating caused the observed human response.

### *D. Human response during use and at assessment visits*

Frontal-plane ankle angle was lower on average across repeated recordings, while individual courses varied and were not consistently monotonic. The recorded target was unchanged between each triggering-session recording and the first subsequent adaptive-session recording, so the paired comparison was not accompanied by a change in target level. However, these pairs came from only seven participants, some adjacent pairs shared recordings, and selecting observations after failure allows a possible return toward usual performance. The contribution of these factors was not quantified.

Evidence for the paired frontal-plane ankle angle change also depends on the uncertainty method: the $t$-based 95% interval with 6 degrees of freedom included zero, whereas the participant-level cluster-bootstrap interval did not. The mixed model produced a boundary warning for participant variance, adding caution to this exploratory estimate. Together with continued practice, session timing, and target progression elsewhere in the course, these limitations prevent attributing the observed trajectories specifically to model updating.

The formal Post and Retention assessments complement the session trajectories by comparing groups at the assessment visits. At Post, the AI group had a lower baseline-adjusted frontal-plane ankle angle than the sequentially recruited CONTROL group, with a consistent direction across conventional, HC3 heteroscedasticity-consistent, and participant-bootstrap intervals. At Retention, these intervals included zero, leaving persistence of a between-group difference unresolved.

Prior CAI visual-feedback work provides context for targeting frontal-plane ankle angle [6], although differences in randomization and co-interventions limit comparisons with the present study. The observed group difference is specific to frontal-plane ankle angle and does not establish effects on symptoms, function, or reinjury.

### *E. Limitations and unresolved questions*

The small development cohort and the seven AI participants limit the precision and generalizability of the findings. The ten clinical CONTROL participants also contributed to population-model development, so the computational and clinical cohorts are not fully independent. Marked class imbalance and a held-out fold without GOOD cycles constrained evaluation across classes. Offline feature construction may also differ from live processing, so their correspondence remains uncertain.

Clinical groups were recruited sequentially without randomization, limiting causal interpretation of the group comparison. Generalization beyond this cohort, probability calibration and threshold selection, and the contribution of participant-specific updating to sustained biomechanical or functional benefit remain unresolved.

## V. Conclusion

An adaptive TCN-based gait-biofeedback workflow was operated across repeated sessions in people with CAI. The population model discriminated operational angle-derived labels in participants excluded from training, while saved candidate models showed higher mean offline BAD-class F1 with lower mean GOOD specificity than their parent models. The AI group had a lower baseline-adjusted frontal-plane ankle angle at Post than the sequentially recruited CONTROL group; the Retention difference remained uncertain. The study provides an integrated evaluation of repeated gait feedback and offline participant-specific updating; the contribution of updating to the observed biomechanical change remains unresolved.

## Acknowledgment

None.

## Funding and disclosures

This work was supported by the Fall 2025 Dodson Research Grant from the Graduate School at The University of Texas at El Paso (awarded to Veronika Lebisova) and the Pilot Grant Program supported by Research & Innovation at The University of Texas at El Paso (awarded to Jaeyoon (Jason) Kim). The authors declare no conflicts of interest.

## Data and code availability

De-identified data supporting the findings of this study are available from the corresponding author upon reasonable request and subject to applicable institutional and participant-consent restrictions. Code for model training and data analysis is available from the corresponding author upon reasonable request.

## References

[1] J. Hertel and R. O. Corbett, “An Updated Model of Chronic Ankle Instability,” Journal of Athletic Training, vol. 54, no. 6, pp. 572–588, 2019. doi: 10.4085/1062-6050-344-18.

[2] L. Luan, D. Orth, P. Newman, R. Adams, D. El-Ansary, and J. Han, “Do individuals with ankle instability show altered lower extremity kinematics and kinetics during walking? A systematic review and meta-analysis,” Physiotherapy, vol. 125, Art. no. 101420, 2024. doi: 10.1016/j.physio.2024.101420.

[3] K. Monaghan, E. Delahunt, and B. Caulfield, “Ankle function during gait in patients with chronic ankle instability compared to controls,” Clinical Biomechanics, vol. 21, no. 2, pp. 168–174, 2006. doi: 10.1016/j.clinbiomech.2005.09.004.

[4] G. Moisan, C. Mainville, M. Descarreaux, and V. Cantin, “Lower limb biomechanics in individuals with chronic ankle instability during gait: A case-control study,” J. Foot Ankle Res., vol. 14, no. 1, Art. no. 36, 2021, doi: 10.1186/s13047-021-00476-6.

[5] P. A. Gribble et al., “Evidence review for the 2016 International Ankle Consortium consensus statement on the prevalence, impact and long-term consequences of lateral ankle sprains,” Br. J. Sports Med., vol. 50, no. 24, pp. 1496–1505, 2016, doi: 10.1136/bjsports-2016-096189.
[6] R. M. Koldenhoven et al., “Gait biofeedback and impairment-based rehabilitation for chronic ankle instability,” Scand. J. Med. Sci. Sports, vol. 31, no. 1, pp. 193–204, 2021, doi: 10.1111/sms.13823.
[7] C. E. Ortega, D. M. Torp, L. Donovan, J. D. Simpson, L. Forsyth, and R. M. Koldenhoven, “Gait-training interventions for individuals with chronic ankle instability: A systematic review and meta-analysis,” J. Athl. Train., vol. 60, no. 5, pp. 332–351, 2025, doi: 10.4085/1062-6050-0499.23.
[8] S. H. Mousavi, F. Khorramroo, H. Minoonejad, and J. Zwerver, “Effects of biofeedback on biomechanical factors associated with chronic ankle instability: A systematic review with meta-analysis,” BMC Sports Sci. Med. Rehabil., vol. 15, no. 1, Art. no. 168, 2023, doi: 10.1186/s13102-023-00780-7.
[9] L. M. A. van Gelder, A. Barnes, J. S. Wheat, and B. W. Heller, “The use of biofeedback for gait retraining: A mapping review,” Clin. Biomech., vol. 59, pp. 159–166, 2018, doi: 10.1016/j.clinbiomech.2018.09.020.
[10] J. J. Tate and C. E. Milner, “Real-time kinematic, temporospatial, and kinetic biofeedback during gait retraining in patients: A systematic review,” Phys. Ther., vol. 90, no. 8, pp. 1123–1134, 2010, doi: 10.2522/ptj.20080281.
[11] C. Lea, M. D. Flynn, R. Vidal, A. Reiter, and G. D. Hager, “Temporal convolutional networks for action segmentation and detection,” in Proc. IEEE Conf. Comput. Vis. Pattern Recognit. (CVPR), 2017, pp. 156–165. [Online]. Available: https://openaccess.thecvf.com/content_cvpr_2017/html/Lea_Temporal_Convolutional_Networks_CVPR_2017_paper.html
[12] K. Zhang, J. Wang, C. W. de Silva, and C. Fu, “Unsupervised Cross-Subject Adaptation for Predicting Human Locomotion Intent,” IEEE Transactions on Neural Systems and Rehabilitation Engineering, vol. 28, no. 3, pp. 646–657, 2020. doi: 10.1109/TNSRE.2020.2966749.
[13] K. Harshe, J. R. Williams, T. D. Hocking, and Z. F. Lerner, “Predicting Neuromuscular Engagement to Improve Gait Training With a Robotic Ankle Exoskeleton,” IEEE Robotics and Automation Letters, vol. 8, no. 8, pp. 5055–5060, 2023. doi: 10.1109/LRA.2023.3291919.
[14] P. A. Gribble et al., “Selection criteria for patients with chronic ankle instability in controlled research: A position statement of the International Ankle Consortium,” J. Orthop. Sports Phys. Ther., vol. 43, no. 8, pp. 585–591, 2013, doi: 10.2519/jospt.2013.0303.

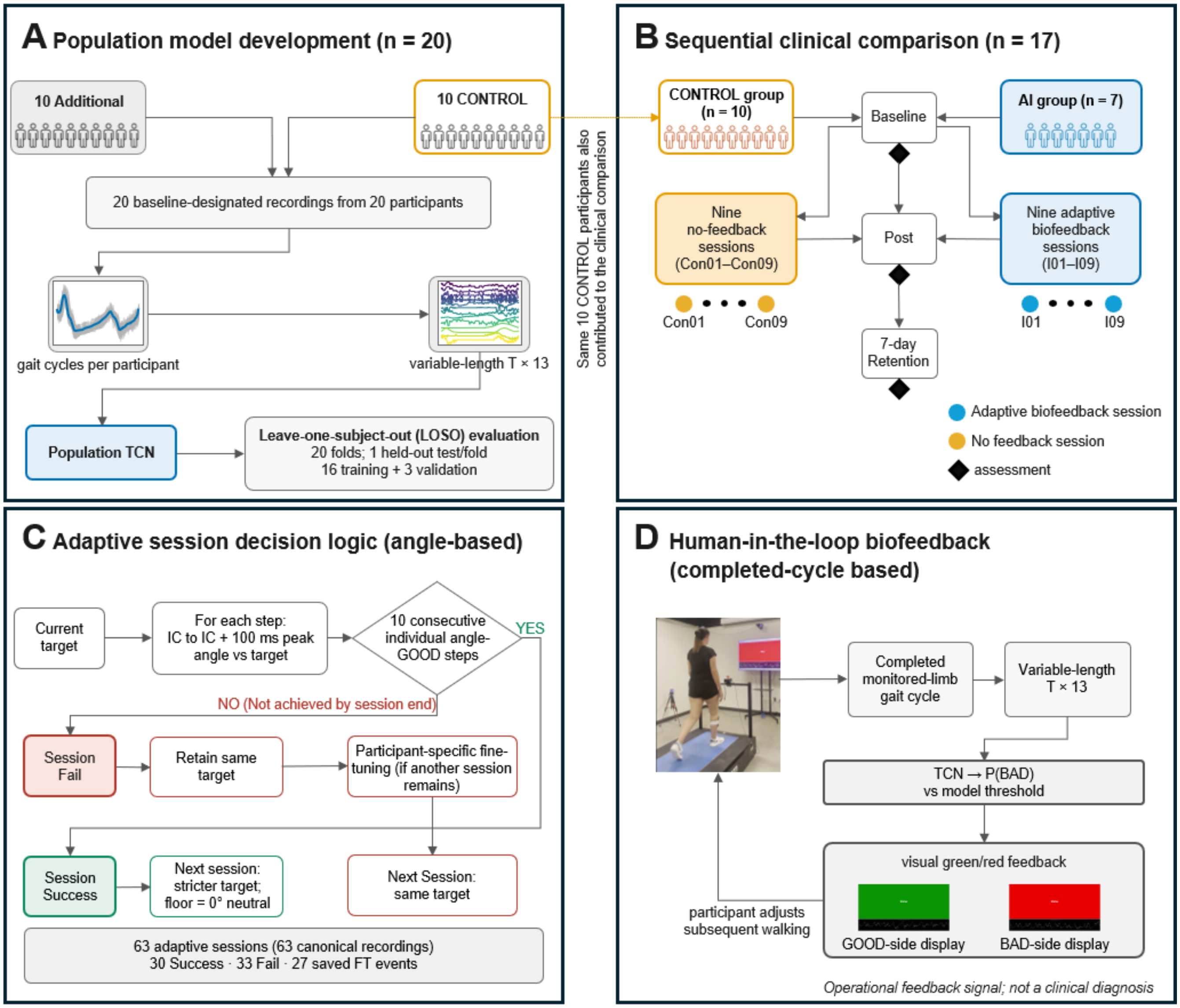

Fig. 1. Study design and biofeedback workflow. (A) Population-model development used baseline-designated recordings from 20 participants, with participant-held-out leave-one-subject-out (LOSO) evaluation. The 10 CONTROL participants also entered the clinical comparison. (B) Both clinical groups completed nine intervention sessions between Baseline and Post: adaptive biofeedback for AI and no-feedback sessions for CONTROL. Baseline, Post, and Retention were separate 5-min assessments without feedback. Post followed the final intervention session, and Retention was assessed seven days after Post. (C) Session progression used each target-meeting step's peak frontal-plane ankle angle within the initial-contact (IC)-to-IC + 100-ms event window. SUCCESS required 10 consecutive target-meeting steps at or below the recorded target. (D) TCN-derived P(BAD) estimates for completed gait cycles drove green/red visual feedback, separately from the angle-based session progression rule. Con01–Con09 denote the nine CONTROL sessions.

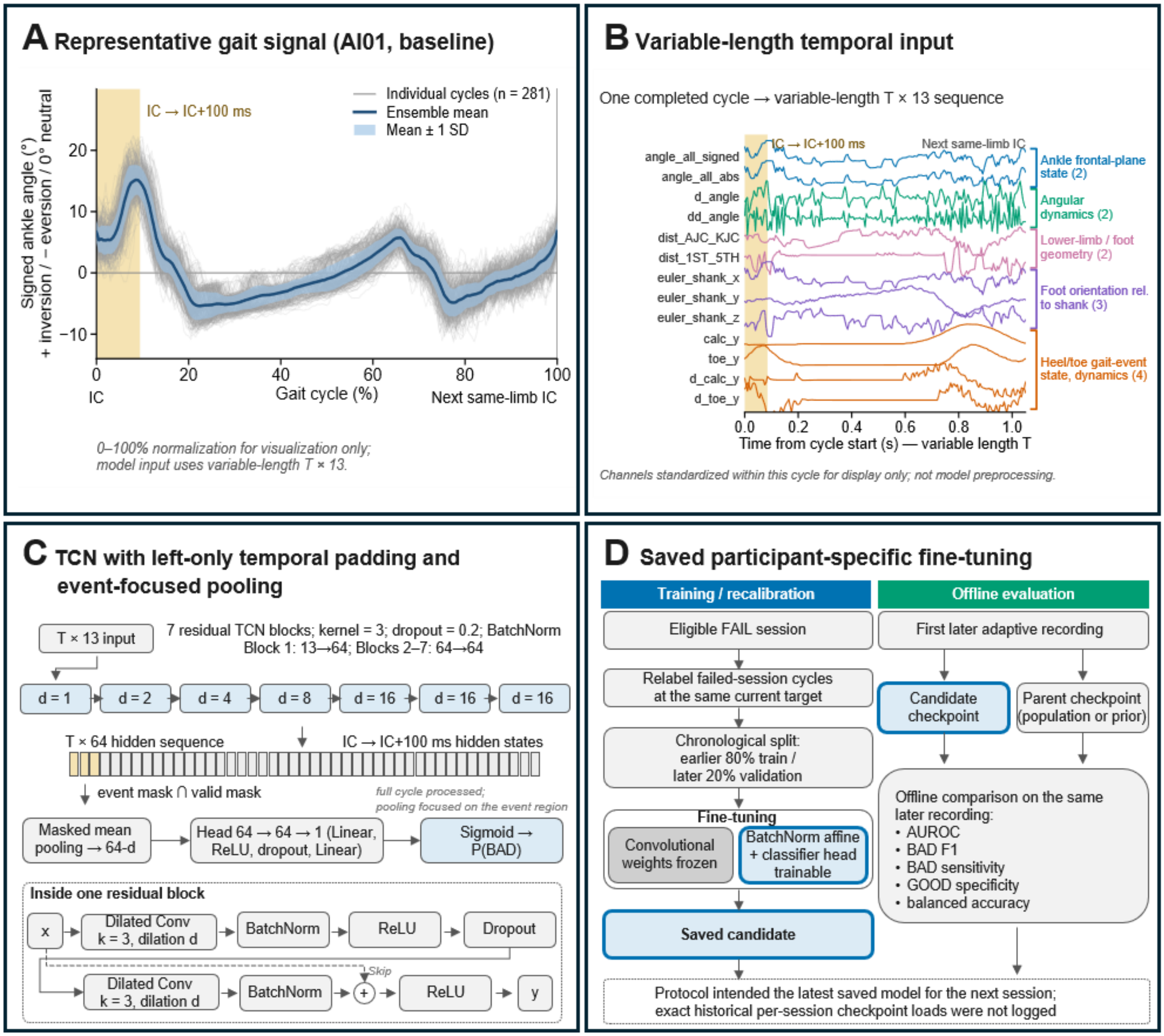

Fig. 2. Model inputs, TCN architecture, and offline evaluation of participant-specific updates. (A) Baseline frontal-plane ankle angle signal from AI01 across 281 cycles. Cycle normalization to 0–100% is used only for display; model inputs retain variable-length T × 13 sequences. (B) Complete 13-channel temporal input representation used by the population model (Full13). (C) TCN with left-padded temporal convolutions and masked mean pooling over the initial-contact (IC)-to-IC + 100-ms event window. (D) Participant-specific updating implemented by fine-tuning, with offline comparison of saved parent and candidate checkpoints on the same-session validation subset and the first subsequent adaptive-session recording.

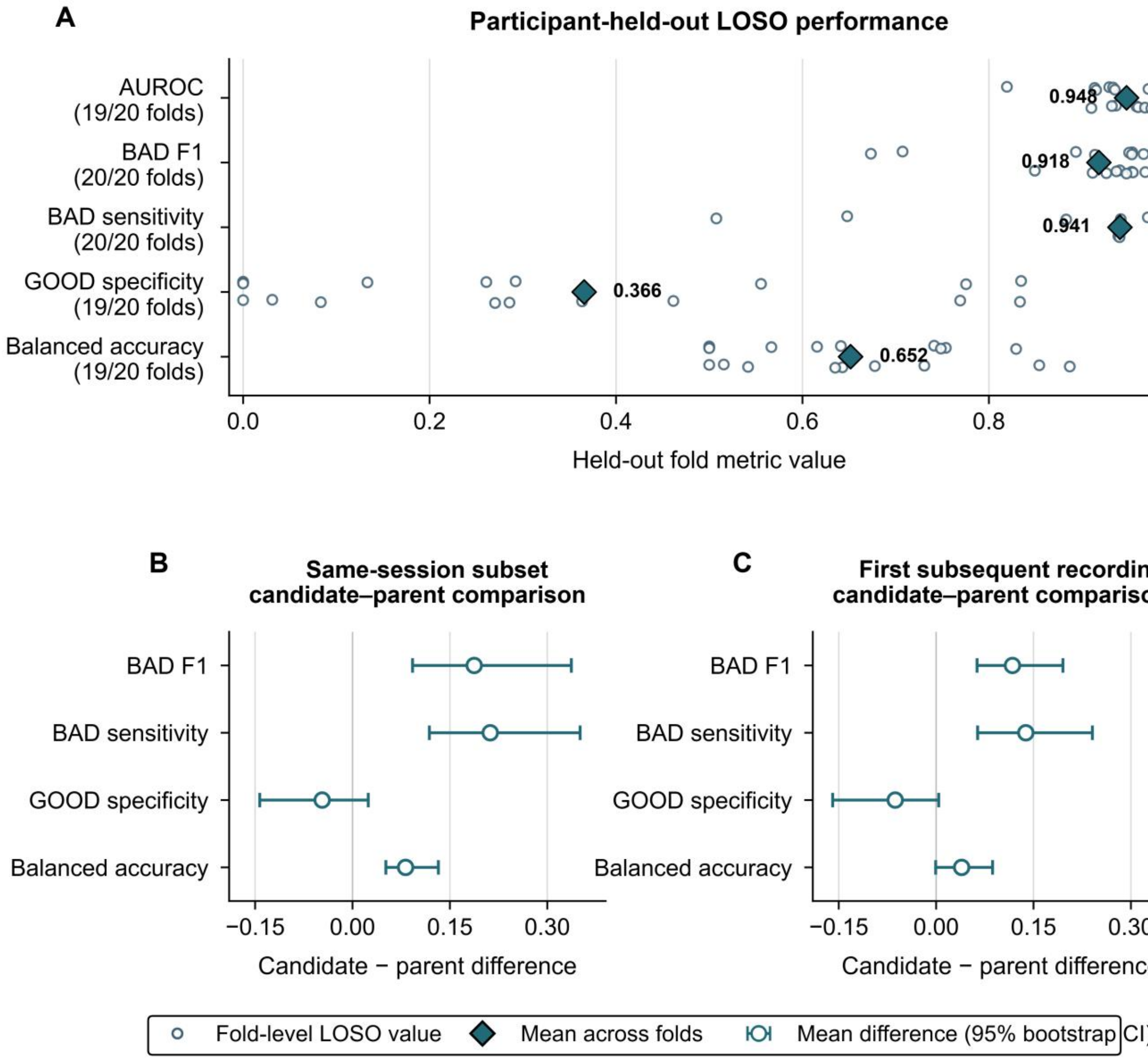

Fig. 3. Population-model performance and offline classification performance after saved model updates. (A) Fold-level LOSO values and means across defined folds for the original population model trained with Full13 inputs. (B) Candidate-minus-parent differences on the same-session validation subset used for candidate selection. (C) Offline candidate-minus-parent differences on the first subsequent adaptive-session recording. Error bars in (B) and (C) show 95% participant-level cluster-bootstrap confidence intervals.

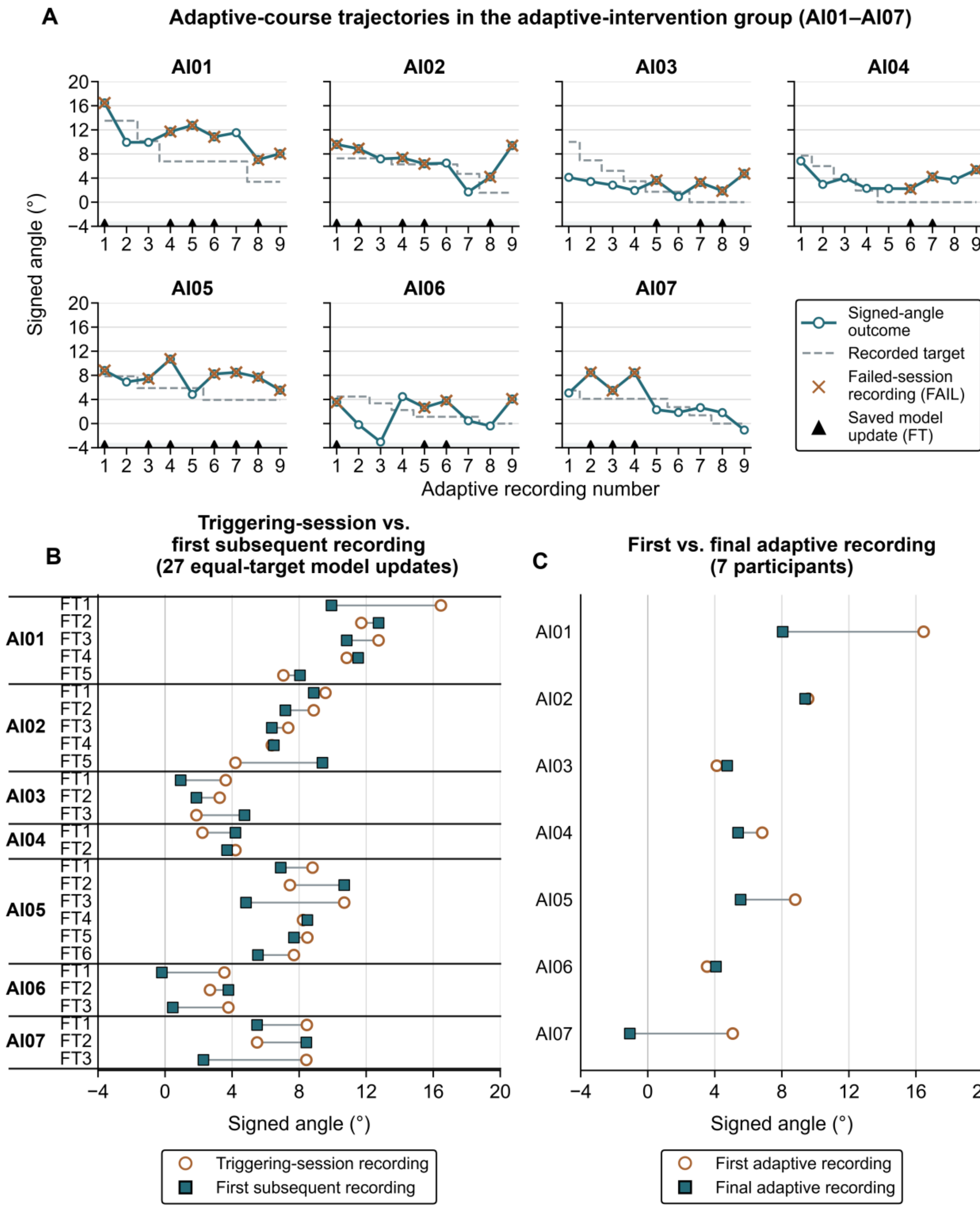

Fig. 4. Frontal-plane ankle angle trajectories during repeated adaptive use. The outcome is the mean of per-step peak frontal-plane ankle angles within the initial-contact (IC)-to-IC + 100-ms event window. SUCCESS required 10 consecutive steps at or below target, so a recording-level mean below target could still coincide with FAIL. (A) Frontal-plane ankle angle outcomes and recorded targets across 63 recordings in seven AI participants (30 SUCCESS and 33 FAIL). Crosses indicate FAIL; triangles indicate the 27 saved model updates. The six failures at the final session were not followed by an update. (B) The triggering-session recording and first subsequent adaptive-session recording for 27 pairs with unchanged targets. (C) First and final adaptive-session recordings. Paired changes are descriptive and cannot isolate the effect of model updating from continued practice.

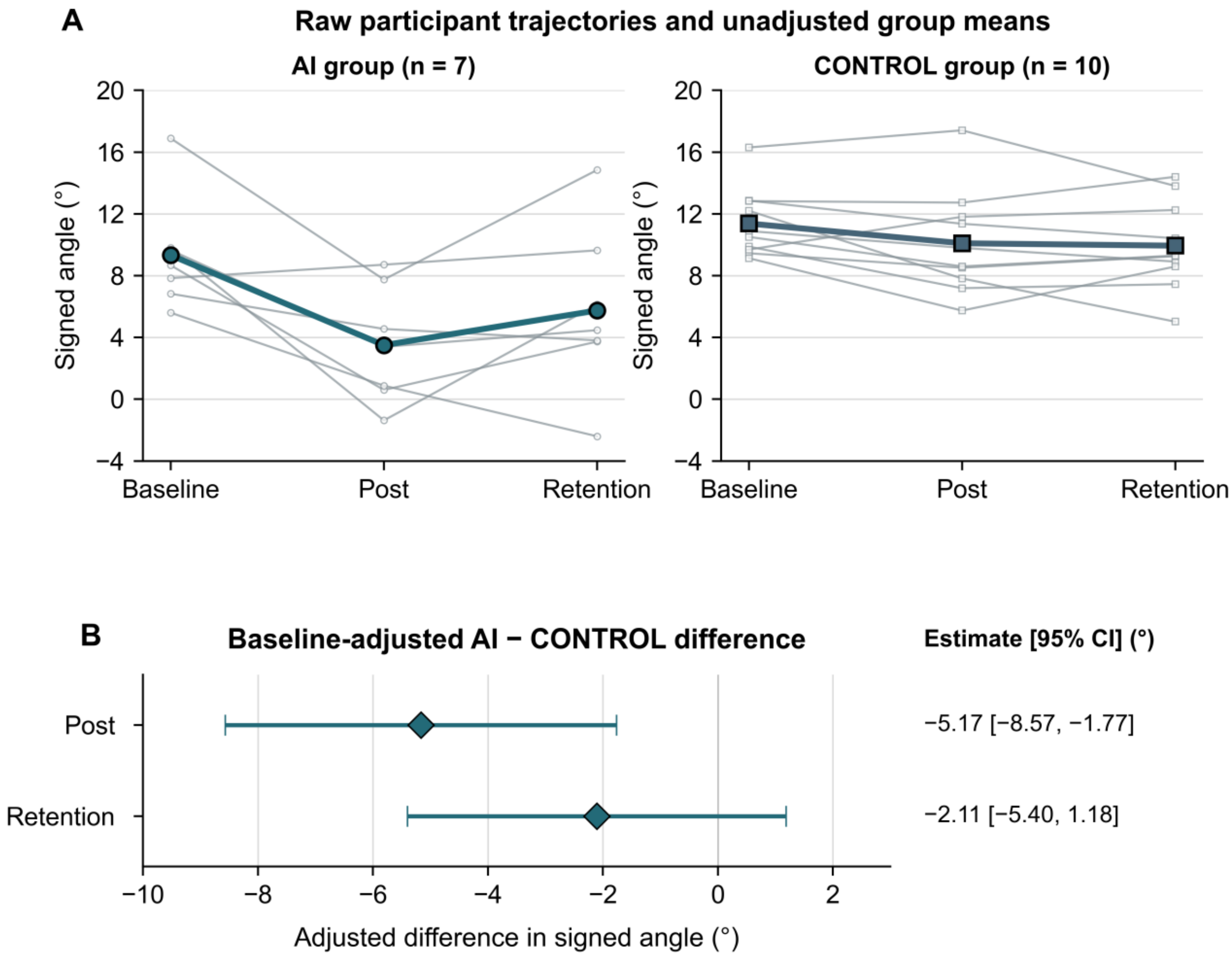

Fig. 5. Frontal-plane ankle angle outcomes in the AI and CONTROL groups. (A) Individual Baseline-to-Post-to-Retention trajectories (thin lines) and unadjusted group means (thick lines). (B) Baseline-adjusted AI-minus-CONTROL differences with 95% confidence intervals (CIs) at Post and Retention.

TABLE I

PARTICIPANT DEMOGRAPHICS AND CLINICAL CHARACTERISTICS

| Characteristic | AI (n = 7)<br>Mean (SD) | CONTROL (n = 10)<br>Mean (SD) |
|---|---|---|
| Sex | 4M, 3F | 4M, 6F |
| Age (years) | 20.0 (0.8) | 23.8 (5.6) |
| Height (cm) | 168.7 (6.8) | 169.4 (10.3) |
| Mass (kg) | 71.0 (14.4) | 87.0 (22.8) |
| Ankle Sprain | 3.4 (2.0) | 2.5 (1.9) |
| Giving Way | 2.6 (1.8) | 4.5 (5.7) |
| Gait Speed (m/s) | 1.13 (0.05) | 1.24 (0.18) |
| IdFAI score | 20.7 (3.7) | 21.3 (4.0) |
| FAAM-ADL (%) | 81.3 (6.8) | 75.3 (7.1) |
| FAAM-Sport (%) | 67.4 (9.3) | 63.1 (8.8) |

Note. Values are mean (SD), except sex, which is reported as counts. M = male; F = female; AI = artificial intelligence-based feedback; IdFAI = Identification of Functional Ankle Instability; FAAM = Foot and Ankle Ability Measure; ADL = activities of daily living.

TABLE II

POPULATION-MODEL PERFORMANCE AND OFFLINE EVALUATION OF PARTICIPANT-SPECIFIC UPDATES

| Metric | Estimate / mean Δ | Defined N | 95% CI | Candidate better |
|---|---|---|---|---|
| **A. Population-model participant-held-out LOSO evaluation** | | | | |
| AUROC | 0.948 | 19/20 folds | — | — |
| BAD F1 | 0.918 | 20/20 folds | — | — |
| BAD sensitivity | 0.941 | 20/20 folds | — | — |
| GOOD specificity | 0.366 | 19/20 folds | — | — |
| Balanced accuracy | 0.652 | 19/20 folds | — | — |
| **B. Same-session evaluation on the candidate-selection validation subset** | | | | |
| Δ BAD F1 | +0.187 | 27/27 events | [+0.093, +0.337] | 22/27 |
| Δ BAD sensitivity | +0.212 | 27/27 events | [+0.118, +0.351] | 16/27 |
| Δ GOOD specificity | -0.047 | 26/27 events | [-0.143, +0.024] | 9/26 |
| Δ Balanced accuracy | +0.082 | 26/27 events | [+0.051, +0.132] | 18/26 |
| **C. Offline evaluation on the first subsequent adaptive-session recording** | | | | |
| Δ BAD F1 | +0.118 | 27/27 events | [+0.063, +0.195] | 23/27 |
| Δ BAD sensitivity | +0.138 | 27/27 events | [+0.064, +0.241] | 13/27 |
| Δ GOOD specificity | -0.063 | 26/27 events | [-0.159, +0.004] | 12/26 |
| Δ Balanced accuracy | +0.039 | 26/27 events | [-0.001, +0.087] | 16/26 |

Note. GOOD/BAD labels were operational gait-cycle labels derived from frontal-plane ankle angle targets. Angle-related signals were model inputs; the recording-specific target was not an explicit population-model input. Parent and candidate denote the saved checkpoints before and after an update, respectively. The same-session comparison used the final 20% validation subset of the triggering-session recording, which also contributed to candidate selection. Both checkpoints were evaluated offline on the first subsequent adaptive-session recording. Defined N gives the number of evaluable folds or events; metrics undefined for recordings containing only BAD cycles were left missing. AUROC = area under the receiver operating characteristic curve; BAD-class F1 = harmonic mean of BAD precision and sensitivity; balanced accuracy = mean of BAD sensitivity and GOOD specificity; CI = confidence interval. Panel A evaluates the population model; panels B and C compare saved model updates offline.

TABLE III

FRONTAL-PLANE ANKLE ANGLE OUTCOMES IN THE AI AND CONTROL GROUPS

| Endpoint | AI n | AI mean ± SD (°) | CONTROL n | CONTROL mean ± SD (°) | Adjusted AI−CONTROL (°) | 95% CI (°) | p |
|---|---|---|---|---|---|---|---|
| Baseline (Descriptive) | 7 | 9.326 ± 3.658 | 10 | 11.374 ± 2.213 | — | — | — |
| Post (Primary) | 7 | 3.501 ± 3.771 | 10 | 10.097 ± 3.373 | -5.168 | [-8.569, -1.766] | 0.0057 |
| Retention (Key Secondary) | 7 | 5.746 ± 5.382 | 10 | 9.941 ± 2.879 | -2.108 | [-5.401, +1.185] | 0.1913 |

Note. Group values are unadjusted participant means ± sample standard deviation (SD). Post was the primary endpoint and Retention the key secondary endpoint. Adjusted AI-minus-CONTROL differences and 95% confidence intervals (CIs) were estimated by analysis of covariance

(ANCOVA) with baseline frontal-plane ankle angle as a covariate. Negative differences indicate lower frontal-plane ankle angles in AI. Robustness analyses are reported in Table S6.

# Supplementary Material

Adaptive gait biofeedback with participant-held-out modeling and participant-specific updating in chronic ankle instability

Tables S1–S7

TABLE S1
COMPUTATIONAL SPECIFICATION OF THE CYCLE CLASSIFIER AND PARTICIPANT-SPECIFIC UPDATING

| Section | Item | Parameter | Value | Detail |
|---|---|---|---|---|
| A — Ordered model inputs | 1 | Frontal-plane ankle angle | deg | Frontal-plane ankle angle relative to neutral (0°); positive values indicate inversion and negative values indicate eversion. Implementation variable: angle_all_signed. |
| A — Ordered model inputs | 2 | Absolute ankle-angle magnitude | deg | Absolute magnitude of the frontal-plane ankle angle. Implementation variable: angle_all_abs. |
| A — Ordered model inputs | 3 | First time derivative of frontal-plane ankle angle | deg/s | First time derivative of the frontal-plane ankle angle. Implementation variable: d_angle. |
| A — Ordered model inputs | 4 | Second time derivative of frontal-plane ankle angle | $deg/s^2$ | Second time derivative of the frontal-plane ankle angle. Implementation variable: dd_angle. |
| A — Ordered model inputs | 5 | Ankle-to-knee joint-center distance | mm | Distance between the ankle joint center (midpoint of the lateral and medial malleolus markers) and the knee joint center (midpoint of the lateral and medial epicondyle markers). Implementation variable: dist_AJC_KJC. |
| A — Ordered model inputs | 6 | First-to-fifth metatarsal marker distance | mm | Distance between the first and fifth metatarsal markers, representing foot width. Implementation variable: dist_1ST_5TH. |
| A — Ordered model inputs | 7 | Foot–shank orientation, ZXY Euler x-component | deg | x-component of the ZXY Euler decomposition of foot orientation relative to the shank; computational frame, not absolute shank orientation. Implementation variable: euler_shank_x. |
| A — Ordered model inputs | 8 | Foot–shank orientation, ZXY Euler y-component | deg | y-component of the ZXY Euler decomposition of foot orientation relative to the shank; computational frame, not absolute shank orientation. Implementation variable: euler_shank_y. |
| A — Ordered model inputs | 9 | Foot–shank orientation, ZXY Euler z-component | deg | z-component of the ZXY Euler decomposition of foot orientation relative to the shank; computational frame, not absolute shank orientation. Implementation variable: euler_shank_z. |
| A — Ordered model inputs | 10 | Calcaneus-marker y-coordinate | mm | y-coordinate of the calcaneus marker in the recorded coordinate system; alignment with |

| Section | Item | Parameter | Value | Detail |
|---|---|---|---|---|
| | | | | anatomical vertical was not verified. Implementation variable: calc_y. |
| A — Ordered model inputs | 11 | Mean metatarsal-marker y-coordinate | mm | Mean y-coordinate of the first and fifth metatarsal markers in the recorded coordinate system; alignment with anatomical vertical was not verified. Implementation variable: toe_y. |
| A — Ordered model inputs | 12 | First time derivative of calcaneus-marker y-coordinate | mm/s | First time derivative of the calcaneus-marker y-coordinate. Implementation variable: d_calc_y. |
| A — Ordered model inputs | 13 | First time derivative of mean metatarsal-marker y-coordinate | mm/s | First time derivative of the mean first/fifth-metatarsal-marker y-coordinate. Implementation variable: d_toe_y. |
| B — TCN architecture | | Input features | 13 | Saved NPZ X second dimension; run_config features channels |
| B — TCN architecture | | Sequence length | variable T | One STEP_WINDOW, not fixed 253-frame crop |
| B — TCN architecture | | Residual blocks | 7 | PretrainModel -> TCNBackbone(num_layers=7) |
| B — TCN architecture | | Hidden channels | 64 | PretrainModel hid=64 and full run_config |
| B — TCN architecture | | Causal convs per block | 2 | ResidualBlock conv1 and conv2 |
| B — TCN architecture | | Kernel size | 3 | PretrainModel TCNBackbone(kernel_size=3) |
| B — TCN architecture | | Dilation schedule | 1;2;4;8;16;16;16 | TCNBackbone default schedule |
| B — TCN architecture | | Convolution padding | left only; (k-1)d per convolution | CausalConv1d F.pad(x,(pad_left,0)), Conv1d padding=0 |
| B — TCN architecture | | Block main path | Conv -> BN -> ReLU -> dropout -> Conv -> BN -> residual add -> ReLU | ResidualBlock forward |
| B — TCN architecture | | Residual projection | 1×1 Conv1d in first block 13->64; identity thereafter | ResidualBlock.down |
| B — TCN architecture | | Dropout | 0.2 | PretrainModel and head default |
| B — TCN architecture | | Backbone output | B×64×T | TCNBackbone forward |
| B — TCN architecture | | Pooling | masked mean over event AND valid masks | masked_mean_pooling; output B×64 |
| B — TCN architecture | | Head | Linear 64->64, ReLU, dropout, Linear 64->1 logits | ClassificationHead |
| B — TCN architecture | | Probability | sigmoid(logit) = P(BAD) | Applied outside head at evaluation/deployment |
| B — TCN architecture | | Total trainable model parameters | 170049 | Analytical parameter count from implemented layer shapes |
| B — TCN architecture | | Personalization variable parameters | 6017 | 14 BN affine pairs = 1792 plus head 4225; conv weights frozen |
| B — TCN architecture | | Theoretical RF slots | 253 frames | Theoretical convolutional receptive field; does not equal a measured amount of available pre-IC history. |
| B — TCN architecture | | Convolutional padding / causality | Left-only temporal padding | Left-only convolutional padding makes the TCN architecture causal conditional on supplied features; centrally differenced |

| Section | Item | Parameter | Value | Detail |
| --- | --- | --- | --- | --- |
| | | | | input features themselves are not established as causal. |
| B — TCN architecture | | Training BN caveat | BatchNorm1d train mode mixes batch/time statistics | run_epoch calls model.train(train_mode), including personalization despite outer BN-eval context |
| B — TCN architecture | | Checkpoint/config | Historical Full13 population run; 20 baseline recordings; 5,475 held-out cycles | run_config.json; checkpoint paths recorded in run metadata |
| C — Training and personalization settings | population LOSO | batch_size | 64 | frozen population LOSO run |
| C — Training and personalization settings | population LOSO | num_epochs | 25 | frozen population LOSO run |
| C — Training and personalization settings | population LOSO | learning_rate | 0.001 | frozen population LOSO run |
| C — Training and personalization settings | population LOSO | eval_split | 0.2 | frozen population LOSO run |
| C — Training and personalization settings | All 27 saved participant-specific FT events | batch_size | 16 | all 27 saved FT run configs independently verified identical |
| C — Training and personalization settings | All 27 saved participant-specific FT events | num_epochs | 15 | all 27 saved FT run configs independently verified identical |
| C — Training and personalization settings | All 27 saved participant-specific FT events | learning_rate | 0.0005 | all 27 saved FT run configs independently verified identical |
| C — Training and personalization settings | All 27 saved participant-specific FT events | patience | 5 | all 27 saved FT run configs independently verified identical |
| C — Training and personalization settings | All 27 saved participant-specific FT events | seed | 42 | all 27 saved FT run configs independently verified identical |
| C — Training and personalization settings | All 27 saved participant-specific FT events | split_mode | contiguous | all 27 saved FT run configs independently verified identical |
| C — Training and personalization settings | All 27 saved participant-specific FT events | eval_split | 0.2 | all 27 saved FT run configs independently verified identical |
| C — Training and personalization settings | All 27 saved participant-specific FT events | save_val_predictions | True | all 27 saved FT run configs independently verified identical |
| C — Training and personalization settings | All 27 saved participant-specific FT events | save_split_manifest | True | all 27 saved FT run configs independently verified identical |
| C — Training and personalization settings | All 27 saved participant-specific FT events | save_label_manifest | True | all 27 saved FT run configs independently verified identical |
| C — Training and personalization settings | evaluation | classification threshold | 0.5 | Frozen evaluation threshold |
| C — Training and personalization settings | personalization | trainability | Convolution frozen; BN affine + classifier head trainable | Verified implementation; BN running-statistics caveat retained |
| C — Training and personalization settings | population labels | recording cutoff | 0.8 × summary Max angle | Historical LOSO run_config labeling rule; summary Max angle is a recording-level mean of step-window peaks; BAD if cycle event-window peak > cutoff, equality GOOD. |
| C — Training and personalization settings | adaptive labels | recorded target | current recording target | Separate participant-specific relabeling; do not substitute population cutoff or nominal target stage. |
| C — Training and personalization settings | population preprocessing | segmentation and gaps | STEP_WINDOW / INV_WINDOW | Historical run_config: skip initial 10 s; cycles from STEP_WINDOW ≥ 1; event mask INV_WINDOW=1; interior marker gaps up to 20 frames linearly interpolated; drop unresolved event-window gaps. |

| Section | Item | Parameter | Value | Detail |
|---|---|---|---|---|
| C — Training and personalization settings | population training | loss and optimizer | BCEWithLogitsLoss; AdamW | Historical run_config training fields; fold validation selects checkpoint, held-out participant remains separate. |
| C — Training and personalization settings | final parent | all-development fit | 20 participants | Separate final Full13 parent checkpoint in pretrain/models/pretrained/full; not one of the LOSO fold models; saved normalization is reused by FT. This does not verify any historical live checkpoint load. |

Note. Section identifies ordered inputs, TCN architecture, or training/updating settings; Item identifies input order or setting group; Parameter gives the reader-facing channel or setting name; Value gives its unit/value; Detail gives its definition and, for model inputs, the implementation variable. TCN = temporal convolutional network; LOSO = leave-one-subject-out; FT = fine-tuning; BN = batch normalization; RF = receptive field; P(BAD) = predicted probability of the BAD operational class. Inputs are ordered and variable length. The population cutoff is 0.8 × the baseline-designated recording's historical IC-window summary Max angle; a cycle is BAD only if its event-window peak exceeds that cutoff (equality GOOD). Adaptive relabeling uses the recorded session target. The 13 channels include four direct angle-derived and nine other kinematic/geometric channels; NoAngle9 removes the former four but is not angle-state independent. Derivative time steps use recorded fps metadata with a 200-Hz software fallback, which is not proof of native acquisition rate. Derivatives used central differences at interior frames. First-derivative boundaries used one-sided differences, whereas second-derivative boundary values copied the nearest interior value. Euler channels represent foot orientation relative to shank. The theoretical 253-frame receptive field does not establish measured pre-IC history. Left-only padding describes TCN convolutions, while centrally differenced inputs are not established causal. Batch-normalization running statistics may update in training mode despite fixed convolutional weights.

TABLE S2

PARTICIPANT-HELD-OUT LOSO FOLD-LEVEL METRICS

| Fold | Held-out participant | Cycles | GOOD | BAD | TP | FP | TN | FN | BAD F1 | BAD sensitivity | GOOD specificity | Balanced accuracy | AUROC |
|---|---|---|---|---|---|---|---|---|---|---|---|---|---|
| 1 | TRAD02 | 264 | 139 | 125 | 81 | 23 | 116 | 44 | 0.707 | 0.648 | 0.835 | 0.741 | 0.819 |
| 2 | TRAD01 | 268 | 65 | 203 | 201 | 46 | 19 | 2 | 0.893 | 0.990 | 0.292 | 0.641 | 0.929 |
| 3 | TRAD03 | 283 | 25 | 258 | 258 | 25 | 0 | 0 | 0.954 | 1.000 | 0.000 | 0.500 | 0.933 |
| 4 | TRAD04 | 260 | 23 | 237 | 230 | 17 | 6 | 7 | 0.950 | 0.970 | 0.261 | 0.616 | 0.913 |
| 5 | TRAD05 | 280 | 15 | 265 | 265 | 13 | 2 | 0 | 0.976 | 1.000 | 0.133 | 0.567 | 0.994 |
| 6 | TRAD06 | 273 | 8 | 265 | 265 | 8 | 0 | 0 | 0.985 | 1.000 | 0.000 | 0.500 | 0.984 |
| 7 | TRAD07 | 289 | 27 | 262 | 133 | 0 | 27 | 129 | 0.673 | 0.508 | 1.000 | 0.754 | 0.971 |
| 8 | TRAD08 | 258 | 17 | 241 | 241 | 17 | 0 | 0 | 0.966 | 1.000 | 0.000 | 0.500 | 0.994 |
| 9 | TRAD09 | 257 | 18 | 239 | 225 | 8 | 10 | 14 | 0.953 | 0.941 | 0.556 | 0.748 | 0.935 |
| 10 | TRAD10 | 271 | 49 | 222 | 196 | 11 | 38 | 26 | 0.914 | 0.883 | 0.776 | 0.829 | 0.915 |
| 11 | CON01 | 277 | 32 | 245 | 245 | 31 | 1 | 0 | 0.940 | 1.000 | 0.031 | 0.516 | 0.936 |
| 12 | CON02 | 313 | 82 | 231 | 231 | 82 | 0 | 0 | 0.849 | 1.000 | 0.000 | 0.500 | 0.932 |
| 13 | CON03 | 272 | 0 | 272 | 272 | 0 | 0 | 0 | 1.000 | 1.000 | NA | NA | NA |
| 14 | CON04 | 294 | 65 | 229 | 215 | 15 | 50 | 14 | 0.937 | 0.939 | 0.769 | 0.854 | 0.958 |
| 15 | CON05 | 253 | 39 | 214 | 214 | 21 | 18 | 0 | 0.953 | 1.000 | 0.462 | 0.731 | 0.988 |
| 16 | CON06 | 264 | 22 | 242 | 240 | 14 | 8 | 2 | 0.968 | 0.992 | 0.364 | 0.678 | 0.960 |
| 17 | CON07 | 276 | 42 | 234 | 220 | 7 | 35 | 14 | 0.954 | 0.940 | 0.833 | 0.887 | 0.967 |
| 18 | CON08 | 274 | 48 | 226 | 226 | 44 | 4 | 0 | 0.911 | 1.000 | 0.083 | 0.542 | 0.910 |
| 19 | CON09 | 268 | 49 | 219 | 219 | 35 | 14 | 0 | 0.926 | 1.000 | 0.286 | 0.643 | 0.974 |
| 20 | CON10 | 281 | 37 | 244 | 244 | 27 | 10 | 0 | 0.948 | 1.000 | 0.270 | 0.635 | 0.991 |

Note. Participant-held-out evaluation used leave-one-subject-out (LOSO) cross-validation. Fold is the LOSO test-fold index; Held-out participant is the participant excluded from training whose cycles form that test fold; Cycles is the number of evaluated cycles; GOOD is the count of operational GOOD, angle-derived labels; BAD is the count of operational BAD, angle-derived labels and the positive operational class. TP counts BAD cycles predicted BAD; FP counts GOOD cycles predicted BAD; TN counts GOOD cycles predicted GOOD; FN counts BAD cycles predicted GOOD. BAD-class F1 is the harmonic mean of BAD precision and BAD sensitivity; BAD sensitivity = TP/(TP+FN); GOOD specificity = TN/(TN+FP); balanced accuracy is the mean of BAD sensitivity and GOOD specificity; AUROC is area under the receiver operating characteristic curve using P(BAD). NA means mathematically undefined, not zero. CON03 has GOOD = 0 and BAD = 272, so its GOOD specificity, balanced accuracy, and AUROC are NA. Metric-specific means average only defined folds: BAD-class F1 and BAD sensitivity are defined in 20/20 folds; AUROC, GOOD specificity, and balanced accuracy in 19/20.

TABLE S3

OFFLINE SAVED-CHECKPOINT PERFORMANCE ON THE FIRST SUBSEQUENT ADAPTIVE-SESSION RECORDING

**A. BAD-class performance**

| Participant | FT event | Triggering-session recording | First subsequent adaptive-session recording | GOOD n | BAD n | Parent BAD F1 | Candidate BAD F1 | Δ BAD F1 | Parent BAD sensitivity | Candidate BAD sensitivity | Δ BAD sensitivity |
|---|---|---|---|---|---|---|---|---|---|---|---|
| AI01 | FT1 | 1 | 2 | 400 | 62 | 0.300 | 0.535 | 0.236 | 1.000 | 0.984 | -0.016 |
| AI01 | FT2 | 4 | 5 | 25 | 1127 | 0.750 | 0.982 | 0.232 | 0.600 | 0.966 | 0.366 |
| AI01 | FT3 | 5 | 6 | 120 | 1070 | 0.947 | 0.978 | 0.031 | 0.901 | 0.966 | 0.065 |
| AI01 | FT4 | 6 | 7 | 131 | 1048 | 0.963 | 0.965 | 0.002 | 0.956 | 0.959 | 0.003 |
| AI01 | FT5 | 8 | 9 | 84 | 1077 | 0.837 | 0.989 | 0.152 | 0.720 | 0.992 | 0.272 |
| AI02 | FT1 | 1 | 2 | 35 | 369 | 0.787 | 0.957 | 0.170 | 0.650 | 0.992 | 0.341 |
| AI02 | FT2 | 2 | 3 | 244 | 376 | 0.825 | 0.833 | 0.008 | 0.989 | 0.987 | -0.003 |
| AI02 | FT3 | 4 | 5 | 355 | 679 | 0.903 | 0.908 | 0.005 | 0.946 | 0.943 | -0.003 |
| AI02 | FT4 | 5 | 6 | 341 | 705 | 0.892 | 0.910 | 0.018 | 0.942 | 0.899 | -0.043 |
| AI02 | FT5 | 8 | 9 | 0 | 1021 | 0.975 | 1.000 | 0.025 | 0.951 | 0.999 | 0.048 |
| AI03 | FT1 | 5 | 6 | 574 | 527 | 0.097 | 0.743 | 0.645 | 0.051 | 0.924 | 0.873 |
| AI03 | FT2 | 7 | 8 | 246 | 867 | 0.807 | 0.855 | 0.048 | 0.757 | 0.754 | -0.002 |
| AI03 | FT3 | 8 | 9 | 36 | 1067 | 0.959 | 0.986 | 0.027 | 0.921 | 0.975 | 0.053 |
| AI04 | FT1 | 6 | 7 | 188 | 968 | 0.439 | 0.975 | 0.536 | 0.281 | 0.988 | 0.707 |
| AI04 | FT2 | 7 | 8 | 137 | 919 | 0.986 | 0.991 | 0.005 | 0.998 | 0.990 | -0.008 |
| AI05 | FT1 | 1 | 2 | 265 | 146 | 0.597 | 0.741 | 0.144 | 1.000 | 0.932 | -0.068 |
| AI05 | FT2 | 3 | 4 | 17 | 816 | 0.979 | 0.994 | 0.015 | 0.960 | 0.994 | 0.034 |
| AI05 | FT3 | 4 | 5 | 641 | 405 | 0.812 | 0.799 | -0.013 | 0.943 | 0.943 | 0.000 |
| AI05 | FT4 | 6 | 7 | 36 | 1024 | 0.964 | 0.987 | 0.023 | 0.933 | 0.983 | 0.051 |
| AI05 | FT5 | 7 | 8 | 137 | 934 | 0.988 | 0.990 | 0.002 | 0.993 | 0.989 | -0.003 |
| AI05 | FT6 | 8 | 9 | 258 | 814 | 0.972 | 0.975 | 0.003 | 0.985 | 0.973 | -0.012 |
| AI06 | FT1 | 1 | 2 | 416 | 6 | 0.000 | 0.000 | 0.000 | 0.000 | 0.000 | 0.000 |
| AI06 | FT2 | 5 | 6 | 46 | 963 | 0.277 | 0.977 | 0.699 | 0.161 | 1.000 | 0.839 |
| AI06 | FT3 | 6 | 7 | 605 | 418 | 0.580 | 0.580 | 0.000 | 1.000 | 1.000 | 0.000 |
| AI07 | FT1 | 2 | 3 | 197 | 474 | 0.692 | 0.959 | 0.266 | 0.530 | 0.954 | 0.424 |
| AI07 | FT2 | 3 | 4 | 217 | 673 | 0.962 | 0.971 | 0.009 | 0.985 | 0.972 | -0.013 |
| AI07 | FT3 | 4 | 5 | 634 | 488 | 0.867 | 0.752 | -0.115 | 0.779 | 0.602 | -0.176 |

**B. Class-balance performance**

| Participant | FT event | Parent GOOD specificity | Candidate GOOD specificity | Δ GOOD specificity | Parent balanced accuracy | Candidate balanced accuracy | Δ balanced accuracy |
|---|---|---|---|---|---|---|---|
| AI01 | FT1 | 0.275 | 0.738 | 0.463 | 0.637 | 0.861 | 0.223 |
| AI01 | FT2 | 1.000 | 0.920 | -0.080 | 0.800 | 0.943 | 0.143 |
| AI01 | FT3 | 0.983 | 0.917 | -0.067 | 0.942 | 0.942 | -0.001 |
| AI01 | FT4 | 0.763 | 0.779 | 0.015 | 0.860 | 0.869 | 0.009 |

| Participant | FT event | Parent GOOD specificity | Candidate GOOD specificity | Δ GOOD specificity | Parent balanced accuracy | Candidate balanced accuracy | Δ balanced accuracy |
|---|---|---|---|---|---|---|---|
| AI01 | FT5 | 1.000 | 0.821 | -0.179 | 0.860 | 0.907 | 0.047 |
| AI02 | FT1 | 0.971 | 0.143 | -0.829 | 0.811 | 0.567 | -0.244 |
| AI02 | FT2 | 0.369 | 0.410 | 0.041 | 0.679 | 0.698 | 0.019 |
| AI02 | FT3 | 0.715 | 0.744 | 0.028 | 0.831 | 0.843 | 0.013 |
| AI02 | FT4 | 0.651 | 0.842 | 0.191 | 0.796 | 0.870 | 0.074 |
| AI02 | FT5 | NA | NA | NA | NA | NA | NA |
| AI03 | FT1 | 1.000 | 0.483 | -0.517 | 0.526 | 0.703 | 0.178 |
| AI03 | FT2 | 0.585 | 0.963 | 0.378 | 0.671 | 0.859 | 0.188 |
| AI03 | FT3 | 1.000 | 0.944 | -0.056 | 0.961 | 0.960 | -0.001 |
| AI04 | FT1 | 1.000 | 0.798 | -0.202 | 0.640 | 0.893 | 0.252 |
| AI04 | FT2 | 0.825 | 0.949 | 0.124 | 0.911 | 0.970 | 0.058 |
| AI05 | FT1 | 0.257 | 0.679 | 0.423 | 0.628 | 0.805 | 0.177 |
| AI05 | FT2 | 1.000 | 0.706 | -0.294 | 0.980 | 0.850 | -0.130 |
| AI05 | FT3 | 0.760 | 0.736 | -0.023 | 0.851 | 0.840 | -0.012 |
| AI05 | FT4 | 0.944 | 0.750 | -0.194 | 0.939 | 0.867 | -0.072 |
| AI05 | FT5 | 0.891 | 0.934 | 0.044 | 0.942 | 0.962 | 0.020 |
| AI05 | FT6 | 0.868 | 0.930 | 0.062 | 0.927 | 0.952 | 0.025 |
| AI06 | FT1 | 1.000 | 1.000 | 0.000 | 0.500 | 0.500 | 0.000 |
| AI06 | FT2 | 1.000 | 0.000 | -1.000 | 0.580 | 0.500 | -0.080 |
| AI06 | FT3 | 0.000 | 0.000 | 0.000 | 0.500 | 0.500 | 0.000 |
| AI07 | FT1 | 1.000 | 0.914 | -0.086 | 0.765 | 0.934 | 0.169 |
| AI07 | FT2 | 0.802 | 0.908 | 0.106 | 0.893 | 0.940 | 0.046 |
| AI07 | FT3 | 0.986 | 1.000 | 0.014 | 0.882 | 0.801 | -0.081 |

Note. Section A: Participant is the publication AI identifier; FT event is the ordered saved participant-specific fine-tuning event; Triggering-session recording is the failed-session recording that initiated it; First subsequent adaptive-session recording is the temporally later dataset used for offline evaluation; GOOD n and BAD n are the operational-label counts in that dataset. Parent BAD-class F1 is the BAD-class F1 from the saved checkpoint immediately before fine-tuning; Candidate BAD-class F1 is the value from the saved post-update checkpoint; Δ BAD-class F1 is candidate minus parent. Parent and candidate BAD sensitivity are the corresponding pre- and post-update values; Δ BAD sensitivity is candidate minus parent. Section B links to A by Participant and FT event. Parent and candidate GOOD specificity and balanced accuracy are the corresponding pre- and post-update values; each Δ is candidate minus parent. BAD is the positive operational class; BAD-class F1 is its precision/sensitivity harmonic mean; balanced accuracy averages BAD sensitivity and GOOD specificity. Both checkpoints were evaluated offline on the same first subsequent adaptive-session recording. NA means a metric is undefined because its required class is absent, not zero. The first subsequent recording is separate in time from the triggering-session recording, but the comparison remains offline and does not establish a causal human response to updating.

TABLE S4
SAVED PARTICIPANT-SPECIFIC MODEL LINEAGE

| Participant | FT event | Triggering-session recording | Triggering-recording target (°) | Parent checkpoint | Candidate checkpoint | First subsequent adaptive-session recording |
|---|---|---|---|---|---|---|
| AI01 | FT1 | 1 | 13.51 | pretrained_best.pth | ft_v1.pth | 2 |
| AI01 | FT2 | 4 | 6.76 | ft_v1.pth | ft_v2.pth | 5 |
| AI01 | FT3 | 5 | 6.76 | ft_v2.pth | ft_v3.pth | 6 |
| AI01 | FT4 | 6 | 6.76 | ft_v3.pth | ft_v4.pth | 7 |
| AI01 | FT5 | 8 | 3.38 | ft_v4.pth | ft_v5.pth | 9 |
| AI02 | FT1 | 1 | 7.27 | pretrained_best.pth | ft_v1.pth | 2 |
| AI02 | FT2 | 2 | 7.27 | ft_v1.pth | ft_v2.pth | 3 |
| AI02 | FT3 | 4 | 6.27 | ft_v2.pth | ft_v3.pth | 5 |
| AI02 | FT4 | 5 | 6.27 | ft_v3.pth | ft_v4.pth | 6 |
| AI02 | FT5 | 8 | 1.57 | ft_v4.pth | ft_v5.pth | 9 |
| AI03 | FT1 | 5 | 1.73 | pretrained_best.pth | ft_v1.pth | 6 |
| AI03 | FT2 | 7 | 0.00 | ft_v1.pth | ft_v2.pth | 8 |
| AI03 | FT3 | 8 | 0.00 | ft_v2.pth | ft_v3.pth | 9 |
| AI04 | FT1 | 6 | 0.00 | pretrained_best.pth | ft_v1.pth | 7 |
| AI04 | FT2 | 7 | 0.00 | ft_v1.pth | ft_v2.pth | 8 |
| AI05 | FT1 | 1 | 7.84 | pretrained_best.pth | ft_v1.pth | 2 |
| AI05 | FT2 | 3 | 5.88 | ft_v1.pth | ft_v2.pth | 4 |
| AI05 | FT3 | 4 | 5.88 | ft_v2.pth | ft_v3.pth | 5 |
| AI05 | FT4 | 6 | 3.92 | ft_v3.pth | ft_v4.pth | 7 |
| AI05 | FT5 | 7 | 3.92 | ft_v4.pth | ft_v5.pth | 8 |
| AI05 | FT6 | 8 | 3.92 | ft_v5.pth | ft_v6.pth | 9 |
| AI06 | FT1 | 1 | 4.48 | pretrained_best.pth | ft_v1.pth | 2 |
| AI06 | FT2 | 5 | 1.12 | ft_v1.pth | ft_v2.pth | 6 |
| AI06 | FT3 | 6 | 1.12 | ft_v2.pth | ft_v3.pth | 7 |
| AI07 | FT1 | 2 | 4.10 | pretrained_best.pth | ft_v1.pth | 3 |
| AI07 | FT2 | 3 | 4.10 | ft_v1.pth | ft_v2.pth | 4 |
| AI07 | FT3 | 4 | 4.10 | ft_v2.pth | ft_v3.pth | 5 |

Note. Participant is the publication AI identifier; FT event is the ordered saved fine-tuning event; Triggering-session recording is the failed-session recording that triggered it; Triggering-recording target (°) is the recorded session target used for participant-specific relabeling, in degrees, and was not an explicit population-TCN input. Parent checkpoint is the saved checkpoint immediately before fine-tuning; Candidate checkpoint is the saved checkpoint produced by that event; First subsequent adaptive-session recording is the temporally later dataset used for offline parent-versus-candidate evaluation. The exact .pth names document the saved model lineage.

TABLE S5
RECORDED TARGET AND RESULT HISTORY FOR 63 CANONICAL ADAPTIVE RECORDINGS

| Participant | Adaptive recording | Recorded target (°) | Frontal-plane ankle angle outcome (°) | Session result | Saved model update after recording |
|---|---|---|---|---|---|
| AI01 | 1 | 13.51 | 16.46 | FAIL | FT1 |
| AI01 | 2 | 13.51 | 9.93 | SUCCESS | — |
| AI01 | 3 | 10.13 | 9.93 | SUCCESS | — |
| AI01 | 4 | 6.76 | 11.71 | FAIL | FT2 |
| AI01 | 5 | 6.76 | 12.74 | FAIL | FT3 |
| AI01 | 6 | 6.76 | 10.84 | FAIL | FT4 |
| AI01 | 7 | 6.76 | 11.52 | SUCCESS | — |
| AI01 | 8 | 3.38 | 7.06 | FAIL | FT5 |
| AI01 | 9 | 3.38 | 8.05 | FAIL | — |
| AI02 | 1 | 7.27 | 9.57 | FAIL | FT1 |
| AI02 | 2 | 7.27 | 8.87 | FAIL | FT2 |
| AI02 | 3 | 7.27 | 7.18 | SUCCESS | — |
| AI02 | 4 | 6.27 | 7.35 | FAIL | FT3 |
| AI02 | 5 | 6.27 | 6.37 | FAIL | FT4 |
| AI02 | 6 | 6.27 | 6.49 | SUCCESS | — |
| AI02 | 7 | 4.70 | 1.70 | SUCCESS | — |
| AI02 | 8 | 1.57 | 4.20 | FAIL | FT5 |
| AI02 | 9 | 1.57 | 9.39 | FAIL | — |
| AI03 | 1 | 10.00 | 4.12 | SUCCESS | — |
| AI03 | 2 | 6.94 | 3.42 | SUCCESS | — |
| AI03 | 3 | 5.20 | 2.83 | SUCCESS | — |
| AI03 | 4 | 3.47 | 1.94 | SUCCESS | — |
| AI03 | 5 | 1.73 | 3.62 | FAIL | FT1 |
| AI03 | 6 | 1.73 | 0.93 | SUCCESS | — |
| AI03 | 7 | 0.00 | 3.26 | FAIL | FT2 |
| AI03 | 8 | 0.00 | 1.87 | FAIL | FT3 |
| AI03 | 9 | 0.00 | 4.74 | FAIL | — |
| AI04 | 1 | 7.72 | 6.83 | SUCCESS | — |
| AI04 | 2 | 5.98 | 2.95 | SUCCESS | — |
| AI04 | 3 | 3.86 | 4.02 | SUCCESS | — |
| AI04 | 4 | 1.93 | 2.29 | SUCCESS | — |
| AI04 | 5 | 0.00 | 2.26 | SUCCESS | — |
| AI04 | 6 | 0.00 | 2.23 | FAIL | FT1 |
| AI04 | 7 | 0.00 | 4.20 | FAIL | FT2 |
| AI04 | 8 | 0.00 | 3.70 | SUCCESS | — |
| AI04 | 9 | 0.00 | 5.39 | FAIL | — |
| AI05 | 1 | 7.84 | 8.80 | FAIL | FT1 |
| AI05 | 2 | 7.84 | 6.90 | SUCCESS | — |

| Participant | Adaptive recording | Recorded target (°) | Frontal-plane ankle angle outcome (°) | Session result | Saved model update after recording |
|---|---|---|---|---|---|
| AI05 | 3 | 5.88 | 7.45 | FAIL | FT2 |
| AI05 | 4 | 5.88 | 10.69 | FAIL | FT3 |
| AI05 | 5 | 5.88 | 4.83 | SUCCESS | — |
| AI05 | 6 | 3.92 | 8.23 | FAIL | FT4 |
| AI05 | 7 | 3.92 | 8.49 | FAIL | FT5 |
| AI05 | 8 | 3.92 | 7.69 | FAIL | FT6 |
| AI05 | 9 | 3.92 | 5.54 | FAIL | — |
| AI06 | 1 | 4.48 | 3.54 | FAIL | FT1 |
| AI06 | 2 | 4.48 | -0.19 | SUCCESS | — |
| AI06 | 3 | 3.36 | -3.05 | SUCCESS | — |
| AI06 | 4 | 2.24 | 4.46 | SUCCESS | — |
| AI06 | 5 | 1.12 | 2.68 | FAIL | FT2 |
| AI06 | 6 | 1.12 | 3.78 | FAIL | FT3 |
| AI06 | 7 | 1.12 | 0.46 | SUCCESS | — |
| AI06 | 8 | 0.00 | -0.40 | SUCCESS | — |
| AI06 | 9 | 0.00 | 4.07 | FAIL | — |
| AI07 | 1 | 5.48 | 5.07 | SUCCESS | — |
| AI07 | 2 | 4.10 | 8.46 | FAIL | FT1 |
| AI07 | 3 | 4.10 | 5.49 | FAIL | FT2 |
| AI07 | 4 | 4.10 | 8.43 | FAIL | FT3 |
| AI07 | 5 | 4.10 | 2.29 | SUCCESS | — |
| AI07 | 6 | 2.73 | 1.84 | SUCCESS | — |
| AI07 | 7 | 1.37 | 2.64 | SUCCESS | — |
| AI07 | 8 | 0.00 | 1.81 | SUCCESS | — |
| AI07 | 9 | 0.00 | -1.07 | SUCCESS | — |

Note. Protocol interpretation. Participant is the publication AI identifier; Adaptive recording is the analyzed motion-capture recording from an adaptive session, with one recording analyzed for each of nine sessions per participant; Recorded target (°) is the actual session-specific threshold in degrees, rather than a nominal stage. Frontal-plane ankle angle outcome (°) is the historical live-summary Max angle in degrees: the mean across recorded steps of each step's maximum frontal-plane ankle angle within the initial-contact (IC)-to-IC + 100-ms event window. It is not the single largest frame angle. Session result is the separately recorded protocol SUCCESS or FAIL classification. SUCCESS required 10 consecutive target-meeting steps, each assessed in the event window against the recorded target. Session result cannot be inferred by directly comparing Frontal-plane ankle angle outcome (°) with Recorded target (°); a displayed outcome below target can coexist with FAIL. Saved model update after recording identifies a failed session with a subsequent adaptive session remaining that was followed by a saved fine-tuning (FT) event; FT1, FT2, and later labels identify the ordered update events. An em dash means no update was saved after that recording. Six terminal I09 FAILs had no update because no later adaptive session remained. Actual recorded targets govern this history.

Exploratory analysis summary. All 27 triggering-session/first-subsequent pairs had equal recorded targets. The repeated-event REML subsequent-minus-triggering estimate was −0.847° (SE 0.556°; t-based 95% CI with 6 degrees of freedom [−2.208°, +0.515°]; p = 0.1789), with a participant-variance boundary-fit warning. A separate participant-level cluster bootstrap (3,000 resamples; seed 20260917), with participants resampled as blocks and all within-participant events retained, gave a 95% CI of [−1.459°, −0.134°]. The finalized-CALC sensitivity estimate was −0.793° (t-based 95% CI with 6 degrees of freedom [−2.166°, +0.579°]; p = 0.2070). The exploratory canonical-recording slope was −0.314° per recording (t-based 95% CI with 6 degrees of freedom [−0.576°, −0.053°]). Five of seven participants had a lower final than first frontal-plane ankle angle value; mean final-minus-first was −2.611°. These observations do not identify a causal updating effect or a monotonic course. REML = restricted maximum likelihood; CI = confidence interval.

TABLE S6
POST AND RETENTION ROBUSTNESS AND INFLUENCE ANALYSES

**A. Robustness analyses**

| Endpoint | Method | Estimate (°) | 95% CI (°) | p |
|---|---|---|---|---|
| Post (Primary) | conventional OLS/ANCOVA | -5.168 | [-8.569, -1.766] | 0.0057 |
| Post (Primary) | HC3 robust | -5.168 | [-8.837, -1.499] | 0.0092 |
| Post (Primary) | participant bootstrap percentile | -5.168 | [-8.151, -1.592] | NA |
| Post (Primary) | finalized CALC sensitivity | -4.807 | [-8.264, -1.351] | 0.0099 |
| Post (Primary) | CON10 earlier Baseline sensitivity | -5.284 | [-8.532, -2.036] | 0.0036 |
| Retention (Key Secondary) | conventional OLS/ANCOVA | -2.108 | [-5.401, 1.185] | 0.1913 |
| Retention (Key Secondary) | HC3 robust | -2.108 | [-5.580, 1.363] | 0.2137 |
| Retention (Key Secondary) | participant bootstrap percentile | -2.108 | [-4.908, 0.455] | NA |
| Retention (Key Secondary) | finalized CALC sensitivity | -1.928 | [-5.344, 1.489] | 0.2463 |
| Retention (Key Secondary) | CON10 earlier Baseline sensitivity | -2.219 | [-5.098, 0.661] | 0.1206 |

**B. Leave-one-out sensitivity**

| Endpoint | Omitted participant | Estimate (°) | 95% CI (°) | p |
|---|---|---|---|---|
| Post (Primary) | AI01 | -4.478 | [-8.999, 0.043] | 0.0519 |
| Post (Primary) | AI02 | -6.048 | [-8.966, -3.131] | 0.0006 |
| Post (Primary) | AI03 | -4.790 | [-8.371, -1.208] | 0.0127 |
| Post (Primary) | AI04 | -5.109 | [-8.824, -1.395] | 0.0108 |
| Post (Primary) | AI05 | -4.251 | [-7.478, -1.025] | 0.0137 |
| Post (Primary) | AI06 | -5.165 | [-8.777, -1.553] | 0.0086 |
| Post (Primary) | AI07 | -5.519 | [-9.016, -2.023] | 0.0047 |
| Post (Primary) | CON01 | -5.148 | [-8.744, -1.552] | 0.0086 |
| Post (Primary) | CON02 | -5.163 | [-8.801, -1.525] | 0.0090 |
| Post (Primary) | CON03 | -5.064 | [-8.311, -1.817] | 0.0050 |
| Post (Primary) | CON04 | -5.205 | [-8.885, -1.525] | 0.0092 |
| Post (Primary) | CON05 | -5.028 | [-8.584, -1.473] | 0.0092 |
| Post (Primary) | CON06 | -5.431 | [-9.041, -1.822] | 0.0063 |
| Post (Primary) | CON07 | -4.747 | [-8.281, -1.214] | 0.0123 |
| Post (Primary) | CON08 | -5.613 | [-9.172, -2.054] | 0.0047 |
| Post (Primary) | CON09 | -5.282 | [-8.917, -1.647] | 0.0078 |
| Post (Primary) | CON10 | -5.443 | [-8.921, -1.966] | 0.0049 |
| Retention (Key Secondary) | AI01 | -3.043 | [-7.375, 1.290] | 0.1531 |
| Retention (Key Secondary) | AI02 | -2.871 | [-5.854, 0.111] | 0.0579 |
| Retention (Key Secondary) | AI03 | -1.900 | [-5.435, 1.635] | 0.2665 |
| Retention (Key Secondary) | AI04 | -1.830 | [-5.383, 1.722] | 0.2858 |
| Retention (Key Secondary) | AI05 | -2.091 | [-5.694, 1.512] | 0.2321 |
| Retention (Key Secondary) | AI06 | -1.633 | [-4.736, 1.470] | 0.2761 |
| Retention (Key Secondary) | AI07 | -2.185 | [-5.699, 1.329] | 0.2022 |
| Retention (Key Secondary) | CON01 | -2.197 | [-5.661, 1.267] | 0.1939 |

| Endpoint | Omitted participant | Estimate (°) | 95% CI (°) | p |
|---|---|---|---|---|
| Retention (Key Secondary) | CON02 | -2.173 | [-5.690, 1.344] | 0.2049 |
| Retention (Key Secondary) | CON03 | -2.139 | [-5.557, 1.279] | 0.1994 |
| Retention (Key Secondary) | CON04 | -1.915 | [-5.451, 1.621] | 0.2630 |
| Retention (Key Secondary) | CON05 | -1.852 | [-5.185, 1.481] | 0.2514 |
| Retention (Key Secondary) | CON06 | -2.245 | [-5.779, 1.289] | 0.1931 |
| Retention (Key Secondary) | CON07 | -1.522 | [-4.792, 1.747] | 0.3328 |
| Retention (Key Secondary) | CON08 | -1.959 | [-5.519, 1.601] | 0.2557 |
| Retention (Key Secondary) | CON09 | -2.078 | [-5.609, 1.453] | 0.2259 |
| Retention (Key Secondary) | CON10 | -2.665 | [-5.565, 0.235] | 0.0686 |

**C. Influence diagnostics**

| Endpoint | Participant | OLS residual (°) | Leverage | Cook's D |
|---|---|---|---|---|
| Post (Primary) | AI01 | -1.024 | 0.603 | 0.147 |
| Post (Primary) | AI02 | 6.244 | 0.161 | 0.326 |
| Post (Primary) | AI03 | -2.444 | 0.146 | 0.044 |
| Post (Primary) | AI04 | -0.337 | 0.144 | 0.001 |
| Post (Primary) | AI05 | -5.202 | 0.145 | 0.196 |
| Post (Primary) | AI06 | -0.024 | 0.254 | 0.000 |
| Post (Primary) | AI07 | 2.788 | 0.193 | 0.084 |
| Post (Primary) | CON01 | 0.227 | 0.118 | 0.000 |
| Post (Primary) | CON02 | 0.040 | 0.102 | 0.000 |
| Post (Primary) | CON03 | 3.879 | 0.295 | 0.328 |
| Post (Primary) | CON04 | -0.246 | 0.130 | 0.000 |
| Post (Primary) | CON05 | 1.618 | 0.117 | 0.014 |
| Post (Primary) | CON06 | -1.866 | 0.118 | 0.019 |
| Post (Primary) | CON07 | 2.887 | 0.123 | 0.049 |
| Post (Primary) | CON08 | -2.793 | 0.140 | 0.054 |
| Post (Primary) | CON09 | -0.895 | 0.106 | 0.004 |
| Post (Primary) | CON10 | -2.853 | 0.105 | 0.039 |
| Retention (Key Secondary) | AI01 | 1.387 | 0.603 | 0.288 |
| Retention (Key Secondary) | AI02 | 5.408 | 0.161 | 0.261 |
| Retention (Key Secondary) | AI03 | -1.348 | 0.146 | 0.014 |
| Retention (Key Secondary) | AI04 | -1.606 | 0.144 | 0.020 |
| Retention (Key Secondary) | AI05 | -0.099 | 0.145 | 0.000 |
| Retention (Key Secondary) | AI06 | -4.350 | 0.254 | 0.339 |
| Retention (Key Secondary) | AI07 | 0.607 | 0.193 | 0.004 |
| Retention (Key Secondary) | CON01 | -1.035 | 0.118 | 0.006 |
| Retention (Key Secondary) | CON02 | -0.538 | 0.102 | 0.001 |
| Retention (Key Secondary) | CON03 | -1.150 | 0.295 | 0.031 |
| Retention (Key Secondary) | CON04 | 1.279 | 0.130 | 0.011 |
| Retention (Key Secondary) | CON05 | 2.976 | 0.117 | 0.052 |

| Endpoint | Participant | OLS residual (°) | Leverage | Cook's D |
|---|---|---|---|---|
| Retention (Key Secondary) | CON06 | -0.969 | 0.118 | 0.006 |
| Retention (Key Secondary) | CON07 | 4.025 | 0.123 | 0.101 |
| Retention (Key Secondary) | CON08 | 0.935 | 0.140 | 0.007 |
| Retention (Key Secondary) | CON09 | 0.239 | 0.106 | 0.000 |
| Retention (Key Secondary) | CON10 | -5.763 | 0.105 | 0.171 |

Note. Section A: Endpoint is Post (Primary) or Retention (Key Secondary); Method identifies conventional ordinary-least-squares analysis of covariance (OLS/ANCOVA), HC3 heteroscedasticity-consistent inference, participant-bootstrap percentile inference, finalized CALC sensitivity using the mean of per-step peak frontal-plane ankle angles within the IC window, computed from the finalized unfiltered CALC file, or an earlier CON10 Baseline sensitivity. Estimate (°) is the baseline-adjusted AI−CONTROL frontal-plane ankle angle contrast in degrees; 95% CI (°) is its method-specific 95% confidence interval in degrees; p is the corresponding p value. HC3 denotes the HC3 heteroscedasticity-consistent covariance estimator. NA for bootstrap p means no p value was defined, not zero. The earlier CON10 Baseline is an alternate sensitivity value, not the authoritative selected baseline. Section B: Endpoint has the same definition; Omitted participant is the single participant excluded from that fit; Estimate (°), 95% CI (°), and p have the same meanings. Post estimates remained negative in 17/17 leave-one-out analyses, and the CI excluded zero in 16/17; AI01 omission was the exception. Retention estimates remained negative in 17/17, with 0/17 CIs excluding zero. Section C: Endpoint again identifies Post or Retention; Participant is the publication identifier for the diagnostic row; OLS residual (°) is the raw observed-minus-fitted OLS residual in degrees; Leverage is the corresponding diagonal hat-matrix value; Cook’s D is Cook’s distance. The Retention contrast remains uncertain.

TABLE S7
INPUT-DEPENDENCE SENSITIVITY OF PARTICIPANT-HELD-OUT CLASSIFICATION

**A. Matched training/evaluation input-dependence sensitivity: Full13 versus NoAngle9**

| Metric | Full13 | NoAngle9 | Defined folds |
|---|---|---|---|
| AUROC | 0.943 | 0.866 | 19/20 |
| BAD F1 | 0.913 | 0.884 | 20/20 |
| BAD sensitivity | 0.921 | 0.918 | 20/20 |
| GOOD specificity | 0.462 | 0.261 | 19/20 |
| Balanced accuracy | 0.690 | 0.587 | 19/20 |

**B. Paired Full13 minus NoAngle9 differences**

| Metric | Mean Δ | 95% percentile CI | Defined pairs |
|---|---|---|---|
| AUROC | +0.077 | [+0.035, +0.122] | 19/20 |
| BAD F1 | +0.030 | [-0.001, +0.065] | 20/20 |
| BAD sensitivity | +0.004 | [-0.075, +0.086] | 20/20 |
| GOOD specificity | +0.202 | [-0.015, +0.426] | 19/20 |
| Balanced accuracy | +0.103 | [+0.025, +0.181] | 19/20 |

Note. Metric identifies AUROC (area under the receiver operating characteristic curve), BAD-class F1 (harmonic mean of BAD precision and sensitivity), BAD sensitivity (BAD true-positive rate), GOOD specificity (GOOD true-negative rate), or balanced accuracy (mean of BAD sensitivity and GOOD specificity). BAD is the positive operational class. Section A compares Full13, the 13-input TCN, with NoAngle9 under matched training and evaluation conditions. NoAngle9 uses the same architecture and training policy after removing frontal-plane ankle angle, absolute ankle-angle magnitude, and the first and second time derivatives of frontal-plane ankle angle; Defined folds is the number of folds with that metric out of 20. This Full13 reference is separate from the original main population-model run. Section B Mean Δ is the paired fold mean Full13 minus NoAngle9; 95% percentile CI uses 5,000 shared fold-index resamples (seed 20260921); Defined pairs counts folds with both metrics defined. CON03 is all BAD, so its AUROC, GOOD specificity, and balanced accuracy are NA (undefined, not zero), leaving 19 defined folds/pairs for those metrics; BAD-class F1 and sensitivity have 20. Remaining nine inputs can retain ankle-state information, and this input-dependence sensitivity does not establish TCN necessity, superiority, independent clinical gait recognition, or human benefit.